\documentclass[journal]{IEEEtran}
\ifCLASSINFOpdf
\else
\fi

\usepackage{amssymb}
\usepackage{latexsym}

\usepackage{url}
\usepackage{xcolor}
\usepackage{arydshln}
\usepackage{algorithm2e}
\usepackage{subcaption}
\usepackage{graphicx}
\usepackage{subcaption}
\usepackage{amsmath,amsfonts,amsthm,amssymb,bm}
\usepackage{t1enc}
\usepackage{multirow}

\usepackage{amsmath,amssymb} 

\definecolor{newcolor}{rgb}{.8,.349,.1}

\begin{document}
%
\title{Convolutional Recurrent Predictor:\\ Implicit Representation for\\ Multi-target Filtering and Tracking}
%
%
%

\author{Mehryar~Emambakhsh,~Alessandro~Bay~and~Eduard~Vazquez
\thanks{M.~Emambakhsh was with Cortexica Vision Systems Ltd. and now with Mesirow Financial, London, UK Email: mehryar\_emam@yahoo.com}
\thanks{A.~Bay is with Cortexica Vision Systems Ltd., London, UK Email: alessandro.bay@cortexica.com}
\thanks{E.~Vazquez is with AnyVision, Belfast, UK Email:  eduardov@anyvision.co}}

%
%

\markboth{}%
{Emambakhsh \MakeLowercase{\textit{et al.}}: Convolutional Recurrent Predictor: Implicit Representation for Multi-target Filtering and Tracking}
%



\maketitle

\begin{abstract}
Defining a multi-target motion model, an important step of tracking algorithms, is a challenging task due to various factors, from its theoretical formulation to its computational complexity. Using fixed models (as in several generative Bayesian algorithms, such as Kalman filters) can fail to accurately predict sophisticated target motions. On the other hand, sequential learning of the motion model (for example, using recurrent neural networks) can be computationally complex and difficult due to the variable unknown number of targets.
In this paper, we propose a multi-target filtering and tracking algorithm which learns the motion model, simultaneously for all targets. It does so from an implicitly represented state map and performing spatio-temporal data prediction.  To this end, the multi-target state is modelled over a continuous hypothetical target space, using random finite sets and Gaussian mixture probability hypothesis density formulations. The prediction step is recursively performed using a deep convolutional recurrent neural network with a long short-term memory architecture, which is trained as a regression block, on the fly, over {\emph{probability density difference}} maps. Our approach is evaluated over widely used pedestrian tracking benchmarks, remarkably outperforming state-of-the-art multi-target filtering algorithms, while giving competitive results when compared with other tracking approaches: The proposed approach generates an average 40.40 and 62.29 optimal sub-pattern assignment (OSPA) errors on MOT15 and MOT16/17 datasets, respectively, while producing 62.0\%, 70.0\% and 66.9\% multi-object tracking accuracy (MOTA) on MOT16/17, PNNL Parking Lot and PETS09 pedestrian tracking datasets, respectively, when publicly available detectors are used.
\end{abstract}


%
\IEEEpeerreviewmaketitle

\section{Introduction}
Spatio-temporal data filtering plays a key role in numerous security, remote sensing, surveillance, automation and forecasting algorithms. As one of the most important steps in a sequential filtering task, prediction (estimation) of the state variables provides an important insight about the past, present and future data.
Particularly for a multi-target filtering and tracking (MTFT) problem, the prediction step conveys the past information about the latent state variables and suggests target {\emph{proposals}}. 
Motion models, which are a fundamental part of the Bayesian filtering paradigm, are used to perform this task. Once the proposals have been established, a correction (update) stage is applied over them through the state-to-measurement space mapping.
Kalman filter assumes linear motion models with Gaussian distributions for both prediction and update steps. Using the Taylor series expansion and deterministic approximation of non-Gaussian distributions, non-linearity and non-Gaussian behaviour are addressed by Extended and Unscented Kalman Filters {(EKF, UKF)}, respectively. Using the importance sampling principle, particle filters are also used to estimate the likelihood and posterior densities, addressing non-linearity and non-Gaussian behaviour \cite{Vo:2005,Moratuwage:2014}. 
Mahler proposed random finite sets (RFS) \cite{Mahler:2003}, which provides an encapsulated formulation of multi-target filtering, incorporating clutter densities and detection, survival and birth of target probabilities. To this end, targets and measurements are assumed to form sets with random cardinalities. 
One approach to represent the target state is to use the Probability Hypothesis Density (PHD) maps \cite{Vo:2005,Vo:2006}. 
Vo and Ma proposed Gaussian Mixture PHD (GM-PHD), which propagates the first-order statistical moments to estimate the posterior as a mixture of Gaussians \cite{Vo:2006}.
While GM-PHD is based on Gaussian distributions, a particle filter-based solution is proposed by Sequential Monte Carlo PHD (SMC-PHD) to address non-Gaussian distributions \cite{Vo:2005}. 
Since a large number of particles should be propagated during SMC-PHD, the computational complexity can be high and hence gating might be necessary \cite{Moratuwage:2014}. 
Cardinalised PHD (CPHD) is proposed to also propagate the RFS cardinality over time \cite{Mahler:2007}, while Nagappa {\emph{et al.}} addressed its intractability \cite{Nagappa:2017}. 
The Labelled Multi-Bernoulli Filter (LMB) \cite{Reuter:2014} performs track-to-track association and outperforms previous algorithms in the sense of not relying on high signal to noise ratio (SNR). 
Vo \emph{et al.} proposed Generalized Labelled Multi-Bernoulli (GLMB) as a labelled multi-target filtering \cite{Vo:2014}.

Since the incoming data is usually noisy, cluttered and variable with time, an \emph{a priori} definition of a motion model applicable to all of the targets is not always straightforward. Such inaccuracies on formulating the targets' state-to-state transition functions used by the Bayesian filters will hence cause erroneous predictions. This phenomenon becomes more evident as the complexity of the motion increases.
A robust filtering algorithm should therefore be capable of {\emph{learning}} such (multi-target) motion behaviour, enabling accurate predictions for the following time steps. 
Recently, machine learning has seen the rise of deep learning methods, achieving state-of-the-art results in many fields, from image classification tasks via convolutional neural networks (CNN) \cite{Krizhevsky:2012} to natural language processing via recurrent neural networks (RNNs) \cite{Graves:2013}. 
CNNs can learn the underlying spatial information by sliding \emph{learnable} filters over the input data. 
On the other hand, as a non-linear dynamical system \cite{Bay2016}, RNNs can store and exploit past information through feedback loops. 
The cyclic connections between units allow RNNs to be suitable for predicting temporal sequences, training the network in such a way that its current (state) outputs are used as input for the following steps. Long short-term memory (LSTM) architecture \cite{hochreiter1997} is introduced to resolve the vanishing gradient phenomenon during the training of RNN.

While RNN and CNN networks are capable of learning the temporal and spatial information from the input signals, respectively, their use for multi-target data analysis is not straightforward. Since in an MTFT problem, the number of targets are constantly changing, the motion can not be easily modelled using a network with a fixed architecture (neural networks usually have fixed and predefined number of input and output neurons). One solution is to allocate an LSTM network to each target \cite{Emambakhsh:2019}, which can significantly increase the computational complexity. 

\begin{figure}[!t]
\centering
\includegraphics[width=1\linewidth]{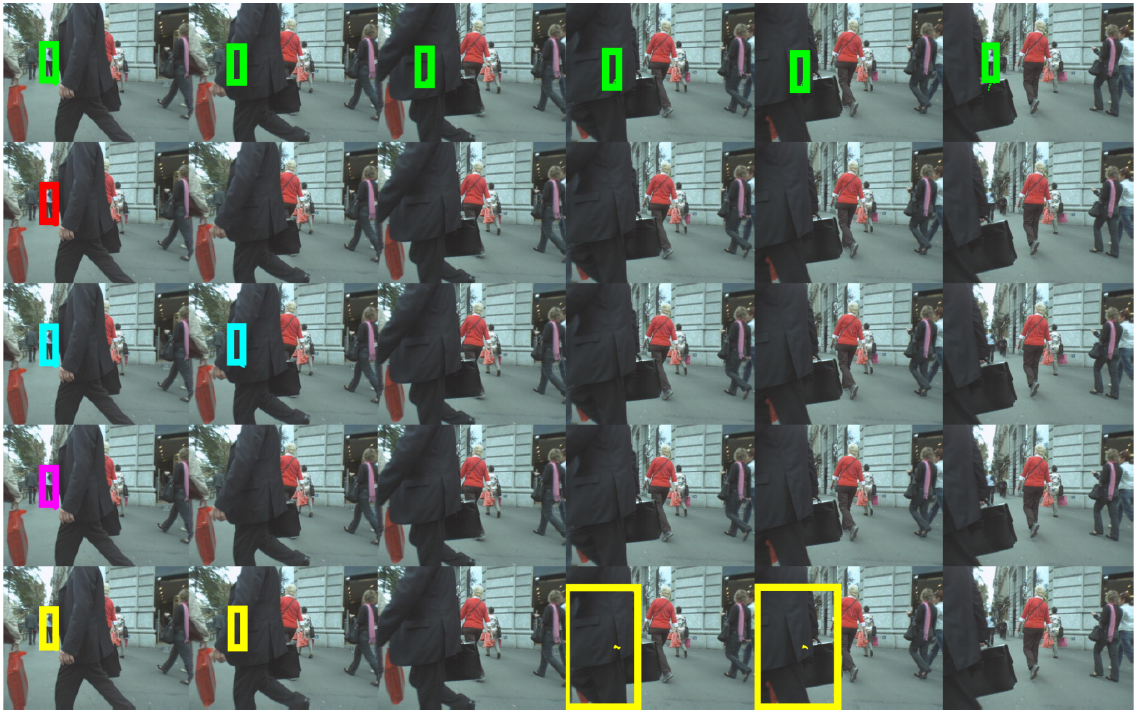}
\caption{Qualitative results: Row 1: our result; Row 2: SORT \cite{Bewley:2016}; Row 3: DeepSORT \cite{wojke2017simple}; Row 4: Re$^3$ \cite{Gordon:2018}; Row 5: RNN{\_}LSTM \cite{Milan:2017}. While the other algorithms lose the occluded target, our proposed MTFT approach maintains its ID until it reappears.}
\label{fig:ReID}
\label{fig:update}
\end{figure}

In this paper, we propose a solution which addresses both of the above problems: (1) the use of fixed models by the Bayesian filtering methods and (2) difficulties in utilising deep neural networks for problems with variable input/output sizes (cardinalities), such as in MTFT. We formulate the MTFT problem from an explicit discrete multi-state vector estimation to an implicit multi-dimensional spatio-temporal prediction. While the former introduces MTFT as a variable (unknown) discrete number of nodes over the state space, the latter performs MTFT implicitly over a continuous state space (such similar explicit vs. implicit intuition exists in Snakes vs. level sets active contours for image segmentation \cite{Emambakhsh:2010}, or $K$-means vs. hierarchical clustering for pattern recognition). Our approach is based on defining probability density difference (PDD) maps, which encapsulate the multi-target information flow over the state space. Inspired by the work proposed by Weinzaepfel \emph{et al.} for dense optical flow computation using convolutional networks \cite{Weinzaepfel:2013}, then, using a network of deep convolutional LSTMs (ConvLSTM \cite{xingjian2015convlstm}) as a regression block, the spatio-temporal prediction is learned and estimated: the spatial dependencies over the multi-target state space are modelled by the convolutional filters, while the temporal dependencies are learned using the LSTM's recurrence. 

As the title suggests, our proposed algorithm is a sequential predictor over probability maps. However, a multi-target Kalman update step over the output prediction maps is also presented, in order to have the paper self-contained. Closely related to the methodology used by an unscented Kalman filter, such update step will be sub-optimal over the non-Gaussian outputs of the ConvLSTM network. This results in a closed-form fast solution, with significantly lower complexity than a similar sequential Monte Carlo (particle filter) approach. It should be mentioned that the use of the Kalman update and its related pre-processing steps are totally arbitrary and other update approaches can also be investigated, which is beyond the scope of this paper.
The algorithm is finally followed by a track-to-track association and target extraction step. 
Our extensive experimental results over several pedestrian tracking benchmarks show remarkable potential of our MTFT algorithm.
Some video samples, further experimental results and notes on mathematical symbols are provided in our Supplementary Material. 

The paper is organised as follows. Section \ref{sec:PDD} introduces the concept of PDD maps, while the whole MTFT pipeline is detailed in Section \ref{sec:overallMTFT}. Experimental results are illustrated in Section \ref{sec:ExpRes} and we conclude the paper in Section \ref{sec:conc}.

{\bf{Scientific contributions:}}
Compared to RFS Bayesian algorithms \cite{Mahler:2007,Nagappa:2017,Reuter:2014,Vo:2014,Vo:2017}, our proposed method models the multi-target motion by learning from the incoming data. The use of the state space, the PDD maps and LSTM networks enable our algorithm to memorise long-term dependencies, as opposed to the detect-to-track tracking methods \cite{Henriques:2015}. To the best of our knowledge, our proposed MTFT algorithm is one of the first methods, which implicitly performs multi-target spatio-temporal prediction by integrating RFS and ConvLSTM. Unlike our previous work \cite{Emambakhsh:2019}, which performs prediction by allocating an LSTM network to each target, our MTFT approach simultaneously estimates state variable for all targets, which significantly increases the computational speed ($\approx 14$ fps). A sample qualitative performance of our algorithm is shown in Fig.~\ref{fig:ReID}.

{\bf{On the mathematical notation:}}
Throughout this paper, we use italic notation for scalars, RFSs, probability density functions (PDFs) and PDD maps. We use bold text for vectors, matrices and tuples. The subscripts and superscripts indicate the time steps for RFSs and scalars/vectors/matrices, respectively. 

\begin{figure*}[!t]
    \centering
    \begin{subfigure}[b]{0.35\textwidth}
        \includegraphics[width=\textwidth]{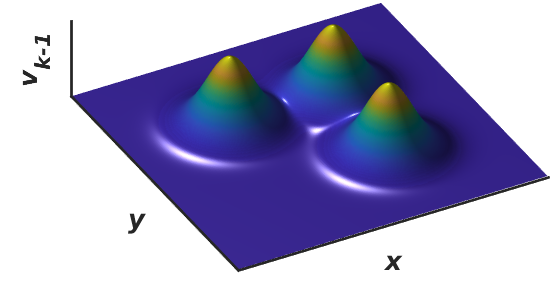}
        \caption{}
    \end{subfigure}
    \begin{subfigure}[b]{0.35\textwidth}
        \includegraphics[width=\textwidth]{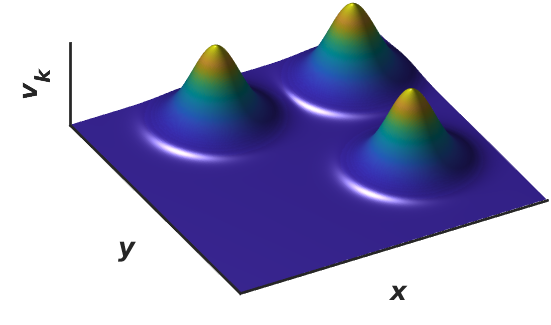}
        \caption{}
    \end{subfigure}
    \begin{subfigure}[b]{0.35\textwidth}
        \includegraphics[width=\textwidth]{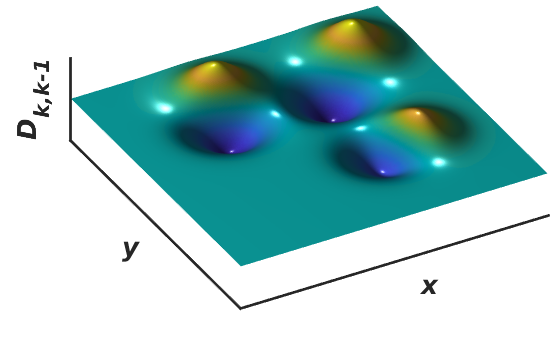}
        \caption{}
    \end{subfigure}
    \caption{An example of PDD generation when: (a) and (b) show the PHD functions $v_{k-1}(x)$ and $v_{k}(x)$ at time steps $k-1$ and $k$, respectively; and (c) illustrates the resulting PDD function $D_{k, k-1}(x) = v_{k}(x) - v_{k-1}(x)$. While, by definition, the multi-target state and cardinality (according to (\ref{eq:PHDCond})) are preserved withing the PHD maps (a) and (b), the PDD map (c) maintains the multi-target state flow (change) over one time step.}
    \label{fig:PDDExplanation}
\end{figure*}


\section{Probability density difference (PDD) maps}
\label{sec:PDD}
Let us define the target state RFS $X_k$ at the $k^{th}$ time step as $X_k = \{{\bf {x}}^{(k)}_1, {\bf {x}}^{(k)}_2, \ldots, {\bf {x}}^{(k)}_{k^\prime}, \ldots, {\bf {x}}^{(k)}_{M_k}\}$, where $M_k$ represents the number of targets (set cardinality). Each ${\bf {x}}^{(k)}_{k^\prime} = \left({\bf{m}}^{(k)}_{k^\prime}, {\bm{\Sigma}}^{(k)}_{k^\prime}, \omega^{(k)}_{k^\prime}, \mathcal{L}^{(k)}_{k^\prime}, a^{(k)}_{k^\prime}, {\bm{\mathcal{M}}}^{(k)}_{k^\prime} \right) \in X_k$ is the ${k^\prime}^{th}$ target state {\emph{tuple}} containing the mean state vector ${\bf{m}}^{(k)}_{k^\prime} \in {\mathbb{R}}^{1 \times d}$ over a $d$-dimensional state space,  covariance matrix ${\bm{\Sigma}}^{(k)}_{k^\prime} \in {\mathbb{R}}^{d \times d}$, Gaussian mixture weight $\omega^{(k)}_{k^\prime} \in {\mathbb{R}}^{1 \times 1}$, integer track label $\mathcal{L}^{(k)}_{k^\prime} \in {\mathbb{Z}}^{1 \times 1}$, target age $a^{(k)}_{k^\prime} \in {\mathbb{Z}}^{1 \times 1}$ (the higher the age, the longer the target has survived) and motion vector ${\bm{\mathcal{M}}}^{(k)}_{k^\prime} \in {\mathbb{R}}^{1 \times 2}$ (along the target's movement direction). 

\begin{figure*}[!t]
\centering
\includegraphics[width=0.8\textwidth]{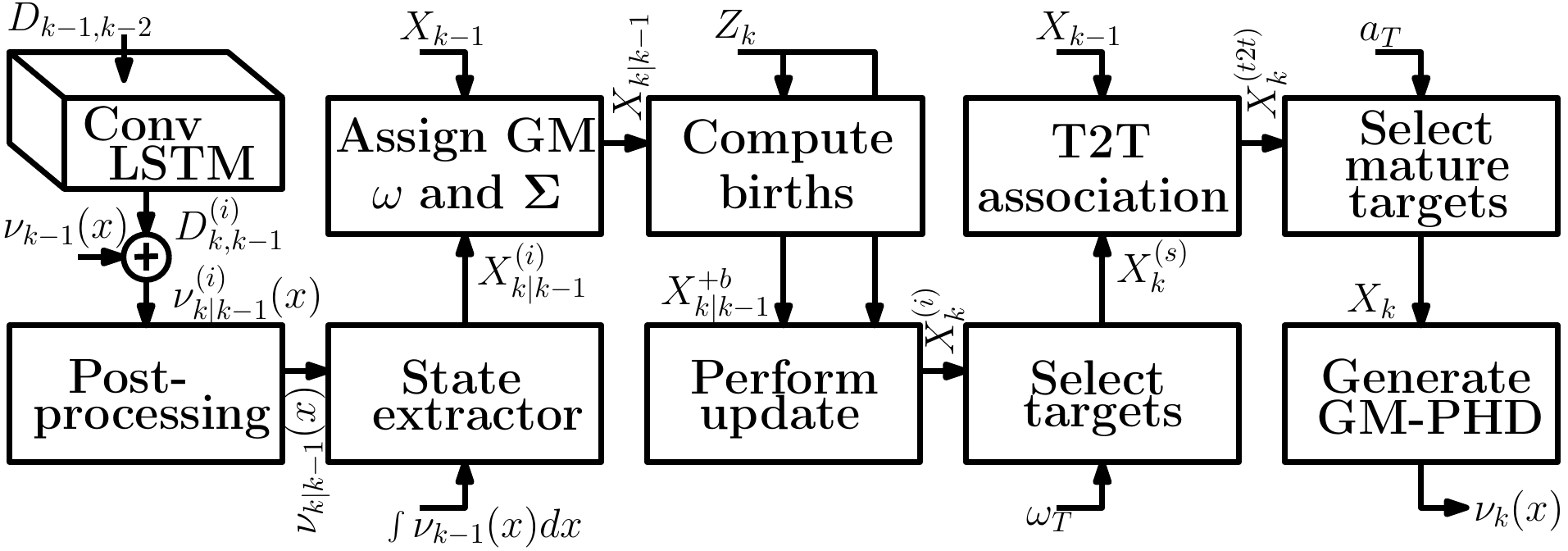}
\caption{Overall MTFT pipeline, transition from $k-1$ to $k$ time step: Once trained, the ConvLSTM network predicts an initial PDD map $D_{k, k-1}^{(i)}$, which is then summed with $v_{k-1}(x)$ to calculate an initial PHD map. Its target tuples are extracted after a post-processing step. We assign initial birth tuples to the measurements at time $k$, then an update step follows. The targets with highest weights are selected which are given to a track-to-track association step for data labelling. Finally, the most mature targets are selected and form a new PHD map $v_k(x)$.}
\label{fig:overall}
\end{figure*}

The target state RFS $X_k$ can be used to create the density function $v_k(x)$ over the (hypothetical) continuous target state $x$, as $v_k(x) = \sum_{k^{\prime}=1}^{M_k} \omega^{(k)}_{k^\prime} \mathcal{N}\left(x| {\bf{m}}^{(k)}_{k^\prime}, {\bm{\Sigma}}^{(k)}_{k^\prime}\right)$,
in which $\mathcal{N}\left(x|{\bf{m}}^{(k)}_{k^\prime}, {\bm{\Sigma}}^{(k)}_{k^\prime}\right)$ is a Normal distribution over $x$, as is assumed in~\cite{Vo:2006}. $v_k(x)$ peaks where the target RFS $X_k$ is located (Fig.~\ref{fig:PDDExplanation}-a and -b). Moreover, $\omega^{(k)}_{k^\prime}$ is assigned such that the following condition is satisfied,
\begin{equation}
\int v_k(x)dx = \sum_{k^\prime=1}^{M_k} \omega^{(k)}_{k^\prime} \approx M_k,
\label{eq:PHDCond}
\end{equation}
which indicates that when $v_k(x)$ is integrated over the target state space $x$, the expected number of targets is given \cite{Mahler:2003,Vo:2006}. When both of these aforementioned properties are satisfied, $v_k(x)$ will represent a GM-PHD function \cite{Vo:2006}.
We define a PDD as the difference between two consecutive GM-PHD functions as follows,
\begin{equation}
D_{k, k-1}\left(x\right) \triangleq v_k(x) - v_{k-1}(x).
\label{eq:pdd}
\end{equation}
\noindent While the PHD function $v_k(x)$ conveys the latent target state information until the $k^{th}$ iteration, the PDD function $D_{k, k-1}\left(x\right)$ contains target state {\emph{flow}} information  between the two consecutive time steps $k-1$ and $k$, emphasizing the most recent variations. The procedure of creating a PDD function from the two consecutive PHD maps is shown in Fig.~\ref{fig:PDDExplanation}. The locations of the peaks in Fig.~\ref{fig:PDDExplanation}-a and -b correspond to the target states (for example, the kinematic attributes of the targets) at time steps $k$ and $k-1$, respectively. Subtracting these two maps creates the PDD map $D_{k, k-1}$ (shown in Fig.~\ref{fig:PDDExplanation}-c), which preserves the transitional information within the multi-target state over one time step.

\section{MTFT pipeline}
\label{sec:overallMTFT}
There is a temporal correlation between a sequence of consecutive PDD maps. Also, assuming a 2D target state, a PDD map can be viewed as a {\emph{texture}} image, in which its pixel values are functions of their location.
The core of our proposed MTFT algorithm is to learn this latent spatio-temporal information within the PDD maps, using a ConvLSTM network: The spatial dependencies between the hypothetical target states are learned using convolutional filters, while the network's recurrence extracts the temporal dependencies. 
To be more specific, we use the ConvLSTM as a spatio-temporal regression block, predicting the next PDD at every time step. Using this approach, (both linear or non-linear) motions are learned by the network, simultaneously for all targets. 
The overall pipeline of our proposed MTFT algorithm is illustrated in Fig.~\ref{fig:overall} for one time step and explained in details in the following sections. First the ConvLSTM network (which is trained on the fly) predicts the target state using the batch of PDD maps. The predicted PDD map is then summed with $v_{k-1}(x)$ to calculate an initial PHD map. The resulting map is then post-processed and its target state tuples are extracted. This is an implicit to explicit representation mapping which allows utilisation of any update algorithms. We assign initial birth tuples to the measurements at time $k$, then an update step follows. Those targets whose GM weights are higher than a given threshold are selected, which are given to a track-to-track association step for label retrieval and/or assignment. Finally, the most mature targets are selected and form a new PHD map.

\begin{figure}[!t]
\centering
\includegraphics[width=0.5\textwidth]{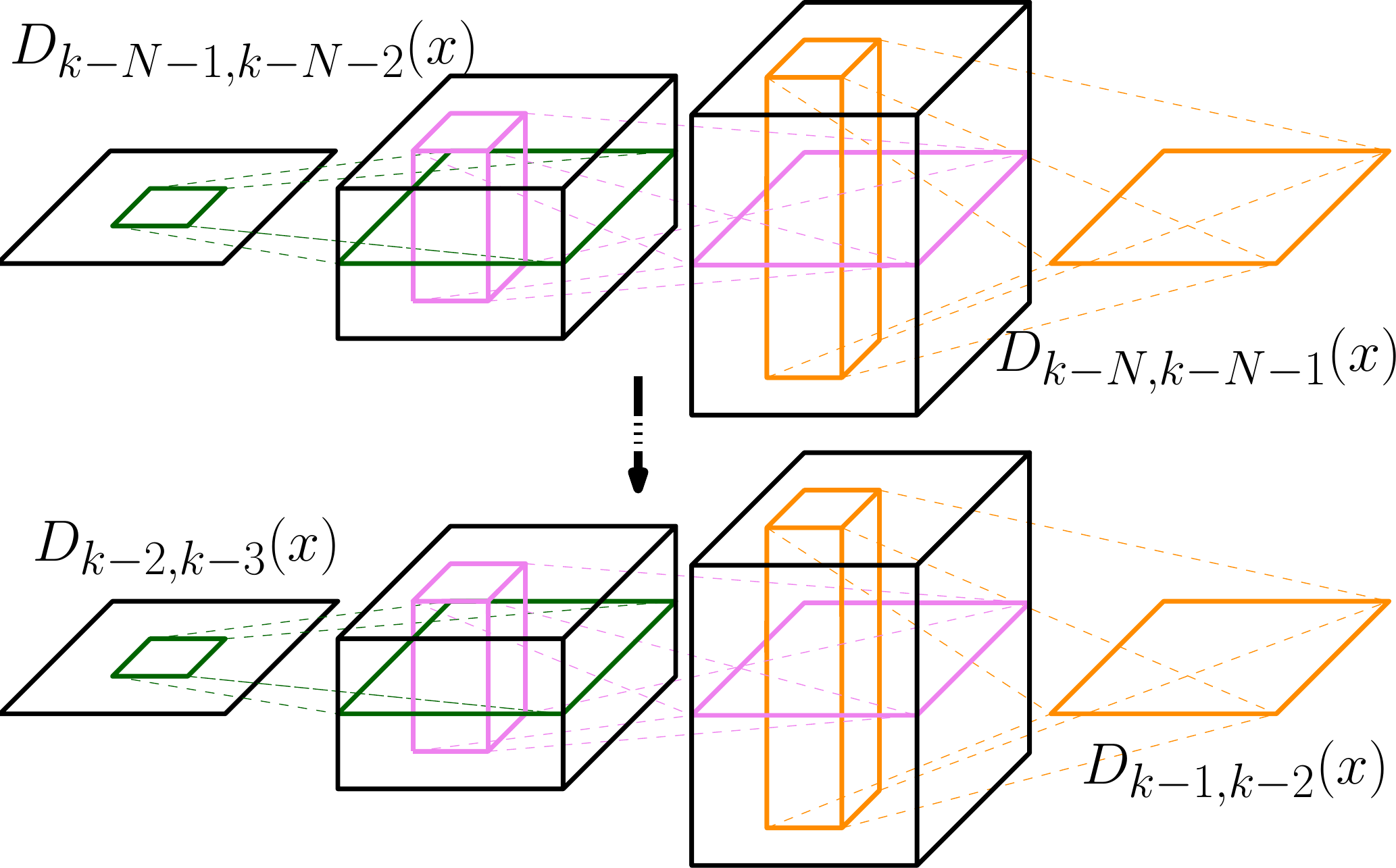}
\caption{The online training step using a ConvLSTM network (with a Many-to-One architecture) over a batch of $N$ PDD maps. Each convolutional filter is shown in different colour. At every epoch during the forward propagation, a batch of previous $N$ PDD maps are used to predict the PDD map at the current time step, which is used to compute the loss function. The back-propagation through time algorithm is then performed to update the network's weights and biases along the steepest descent direction over the objective function.}
\label{fig:pretrain}
\end{figure}

\subsection{Online training step}
The multi-target motion model is learned during the online training step. A batch of $N$ PDD maps are used to train a ConvLSTM model as shown in Fig.~\ref{fig:pretrain}. At every epoch of the training phase, a Kullback-Leibler (KL) divergence loss is calculated between the predicted and true PDD maps during the forward pass and minimised in the back-propagation step.  
KL divergence assumes the output of the ConvLSTM network as PDFs and computes the relative entropy between the prediction and ground truth. The convolutional filters (shown as different colour in Fig.~\ref{fig:pretrain}) learn the spatial information within each PDD map, while the LSTM models the latent temporal state between each filtered PDD map (shown as tensor blocks). The neural network learns how to predict the PDD map for the current time step, given $N$ previous PDD maps.
This trained ConvLSTM network (which is actually a spatio-temporal regression block) is then used to predict the multi-target state map, explained in the next section. 

\subsection{Prediction and implicit to explicit representation}
\label{sec:PredImtoEX}
$D_{k-1, k-2}(x)$ (the output PDD of Fig.~\ref{fig:pretrain}) is given to the trained ConvLSTM network to compute an initial predicted PDD $D^{(i)}_{k, k-1}(x)$ (the input for Fig.~\ref{fig:overall}). When $D^{(i)}_{k, k-1}(x)$ is summed with the PHD function $v_{k-1}(x)$, it returns the (initial) predicted PHD map $v^{(i)}_{k|k-1}(x) = D^{(i)}_{k, k-1}(x) + v_{k-1}(x)$ (see (\ref{eq:pdd})). 
Due to the non-linearity and weights/biases multiplications/additions imposed by the neural network to the input data, the output $v^{(i)}_{k|k-1}$ may not satisfy the PHD conditions (\ref{eq:PHDCond}). Also, because of the padding effect caused by the convolutional filters, there may be artifacts added to the borders of the output data. In order to resolve these issues, the boundary values of $v^{(i)}_{k|k-1}(x)$ are replaced with the median of the inner parts. 
Moreover, we assume that the prediction step does not alter the number of targets. Since the number of targets is equal to the integration of the PHD filter (see (\ref{eq:PHDCond})), after the median filtering is applied, the output map is normalised such that it is integrated to $M_{k-1} =\int v_{k-1}(x)dx$. The result of this post-processing step is the PHD function $v_{k|k-1}(x)$, which is then used to extract the predicted target state RFS $X_{k|k-1}$.
The location of the peaks in a GM-PHD function correspond to the mean vectors of the target states \cite{Vo:2006}. Therefore, in order to extract the explicit target states from $v_{k|k-1}$, first, its peaks are found as follows,
\begin{equation}
\left\{
\begin{array}{l}
{m}_{k|k-1} = \underset{x}{\mathrm{argmax}} \left(v_{k|k-1}(x), M_{k|k-1} \right) \\
{{m}}_{k|k-1} = \left\{{\bf{m}}^{(k|k-1)}_1, {\bf{m}}^{(k|k-1)}_2, \ldots, {\bf{m}}^{(k|k-1)}_{M_{k|k-1}} \right\}
\end{array}
\right.
\label{eq:PeakExtraction}
\end{equation}
where $\underset{x}{\mathrm{argmax}} \left(\bullet, M_{k|k-1} \right)$ computes the $M_{k|k-1}$ highest peaks of the input PHD function. $m_{k|k-1}$ is an RFS with $M_{k|k-1} = M_{k-1} = \int v_{k-1}(x)dx$ cardinality, containing the predicted $d$-dimensional target state mean vectors. The peak values of $v_{k|k-1}(x)$, correspond to the GM-PHD weights which are computed as follows,
\begin{equation}
\left\{
\begin{array}{l}
\Omega_{k|k-1} = \underset{x}{\mathrm{max}} \left(v_{k|k-1}(x), M_{k|k-1} \right)\\
\Omega_{k|k-1} = \left\{\omega^{(k|k-1)}_1, \omega^{(k|k-1)}_2, \ldots, \omega^{(k|k-1)}_{M_{k|k-1}} \right\}
\end{array}
\right.
\end{equation}
where $\Omega_{k|k-1}$ is an RFS containing the GM-PHD peaks for the $M_{k|k-1}$ targets. 
In order to compute the covariance RFS ${\Sigma}_{k|k-1}$, we have examined two approaches. The first one uses ${m}_{k|k-1}$ and $\Omega_{k|k-1}$ as the location and height of a Gaussian mixture, respectively. Then fits a {2D} mixture of Gaussian functions to the PHD map $v_{k|k-1}(x)$ to compute the covariance matrices. 
Another solution is based on finding the corresponding pairs between mean RFS ${{m}}_{k-1}$ and ${{m}}_{k}$ using combinatorial optimisation. ${\Sigma}_{k|k-1}$ is then assigned to its corresponding elements from ${\Sigma}_{k-1}$. We have observed that both of these approaches generate similar results, while the latter is significantly faster, as it is not optimising over a continuous parameter space (unlike the {2D} Gaussian fitting) and is less vulnerable to stop at local minima.
The overall approach explained above can be interpreted as a mapping from an implicit representation ($v_{k|k-1}(x)$) to an explicit target state representation ($X_{k|k-1}$). 

The union of $X_{k|k-1}$ and the birth RFS $X^{(b)}_{k}$, which are assigned using the current measurement RFS $Z_k$, is then computed as follows,
\begin{align}
&\left\{
\begin{array}{l}
X^{(b)}_{k} = \left\{  {{\bf{x}}^{b}_1}^{(k)}, {{\bf{x}}^{b}_2}^{(k)}, \ldots, {{\bf{x}}^{b}_{k^\prime}}^{(k)}, \ldots, {{\bf{x}}^{b}_{M^{b}_k}}^{(k)}    \right\}\\
{{\bf{x}}^{b}_{k^\prime}}^{(k)} = \bigg( {{\bf{m}}_{k^\prime}^b}^{(k)}, {{\bm{\Sigma}}}_{birth}, {\omega}_{birth}, \mathcal{L}_{birth},
\\\hspace{1.5cm} a_{birth}, {\bm{\mathcal{M}}}_{birth} \bigg) \in X^{(b)}_{k}
\end{array}
\right.
\nonumber\\
&X^{(+b)}_{k|k-1} = X_{k|k-1} \cup X^{(b)}_{k} ,
\end{align}
where ${{\bf{x}}^{b}_{k^\prime}}^{(k)}$ is the ${k^\prime}^{th}$ birth target tuple, initialised with covariance matrix ${{\bm{\Sigma}}}_{birth}$, birth weight ${\omega}_{birth}$, birth label identifier $\mathcal{L}_{birth}$, birth age $a_{birth}$ and birth motion vector ${\bm{\mathcal{M}}}_{birth}$.
The predicted RFS $X^{(+b)}_{k|k-1}$ is then updated using the measurement RFS $Z_k$, which is explained in the next section.

\subsection{Update step}
\label{sec:Update}
Assuming ${\bf{z}} \in Z_k$ is a $d_m$-dimensional measurement vector, the updated GM-PHD mean, covariance matrix and Gaussian weights are computed as follows \cite{Vo:2006},
\begin{align}
&{{\bf{m}}^i}^{(k)}_{k^\prime} \left({\bf{z}}\right) = {{\bf{m}}_{k^\prime}^{+b}}^{(k|k-1)} + {\bf{K}}^{(k)}_{k^\prime} \left({\bf{z}} - {\bf{H}}^{(k)} {{{\bf{m}}_{k^\prime}^{+b}}^{(k|k-1)}}^{\intercal} \right)
\nonumber\\
&{{\bm{\Sigma}}^i}^{(k)}_{k^\prime} = \left[{\bf{I}} - {\bf{K}}^{(k)}_{k^\prime} {\bf{H}}^{(k)}\right] {{\bm{\Sigma}}_{k^\prime}^{+b}}^{(k|k-1)}
\\
&{{\omega^{i}}}^{(k)}_{k^{\prime}}\left({\bf{z}}\right) = \frac{p_{D, k}{{\omega^{+b}_{k^{\prime}}}}^{(k|k-1)} q^{(k)}_{k^{\prime}}\left({\bf{z}}\right)}{\kappa_k({\bf{z}}) + p_{D, k} \sum_{k^{\prime\prime} = 1}^{M^{+b}_{k|k-1}}   {{\omega^{+b}_{k^{\prime\prime}}}}^{(k|k-1)} q^{(k)}_{k^{\prime\prime}}\left({\bf{z}}\right)} \nonumber
\label{eq:CovUpdate}
\end{align}
where $\bf{I}$ is a $d \times d$ identity matrix, ${{\bf{m}}_{k^\prime}^{+b}}^{(k|k-1)}$, ${{\bm{\Sigma}}_{k^\prime}^{+b}}^{(k|k-1)}$ and ${{\omega}^{+b}_{k^\prime}}^{(k|k-1)}$ are the $1 \times d$ mean vector, $d \times d$ covariance matrix and $1 \times 1$ Gaussian weight of the ${k^\prime}^{th}$ member of $X^{(+b)}_{k|k-1}$, respectively. ${\bf{H}}^{(k)}$ is a $d_m \times d$ prediction to measurement space mapping matrix. $p_{D, k}$ is the probability of detection and $\kappa_k(\bullet)$ is the clutter intensity at time $k$. $q^{(k)}_{k^{\prime\prime}}\left(\bullet\right)$ is a Gaussian distribution, over the measurement space at time $k$, with updated mean and covariance matrix using the ${{k^{\prime\prime}}}^{th}$ target, i.e.\
\begin{align}
q^{(k)}_{k^{\prime\prime}}\left(\bf{z}\right) =
 \mathcal{N} \bigg( {\bf z}\bigg|&{\bf{H}}^{(k)} {{{\bf{m}}_{k^\prime}^{+b}}^{(k|k-1)}}^{\intercal}, \nonumber\\ &{\bf{R}}^{(k)} + {\bf{H}}^{(k)} {\bm \Sigma}^{(k|k-1)}_{k^{\prime\prime}} {{\bf{H}}^{(k)}}^{\intercal}\bigg)
\end{align}
where ${\bf{R}}^{(k)}$ is a $d_m \times d_m$ covariance of measurement noise. ${\bf{K}}^{(k)}_{k^\prime}$ is a $d \times d_m$ Kalman gain matrix for the ${k^\prime}^{th}$ predicted target computed as:
$$
{\bf{K}}^{(k)}_{k^\prime} = {{\bm{\Sigma}}_{k^\prime}^{+b}}^{(k|k-1)} {{\bf{H}}^{(k)}}^\intercal \left({\bf{H}}^{(k)} {{\bm{\Sigma}}_{k^\prime}^{+b}}^{(k|k-1)} {{\bf{H}}^{(k)}}^\intercal + {\bf{R}}^{(k)} \right)^{-1}
$$
After computing the update state tuples, the Gaussian mixture pruning and merging steps explained in \cite{Vo:2006} are performed over the targets. In order to allow propagating mean state vectors corresponding to sudden target birth, the ``maximum allowable number of Gaussian terms" ($J_{max}$ in \cite{Vo:2006}) is selected by computing the maximum between $M_{k-1}$ (the number of targets in the previous time step) and a sample from a Poisson distribution with $M_{k-1}$ mean. 

\begin{algorithm}[!t]
\KwIn{${\bf{\Delta}}_{k, k-1}, X^{(s)}_k, X^{(t2t)}_k = \{()\}, X_k = \{()\}, X_{k-1},  a_{init}, a_{T}$}
\KwOut{$X_k$}
${\bf{MM}} = \textrm{HM}({\bf{\Delta}}_{k, k-1})$
\% Outputs an $M_{k-1} \times M^{(s)}_{k}$ binary matrix\\
$assigned\_ind = \textrm{RowSum}(\bf{MM}) == 1$\\ 
$unassigned\_ind = \textrm{RowSum}(\bf{MM}) == 0$\\
$unassigned\_prev\_ind = \textrm{ColSum}(\bf{MM}) == 0$\\
\For{$i$ in $assigned\_ind$}{
\% ${{\bf{x}}_i^{(s)}}^{(k)} \in X^{(s)}_k$ survives:\\
Increment age, keep the label and append to $X^{(t2t)}_k$, update motion vector with the associated track;
}
\For{$i$ in $unassigned\_ind$}{
\% ${{\bf{x}}_i^{(s)}}^{(k)} \in X^{(s)}_k$ is a birth:\\
Assign $a_{birth}$, $\mathcal{L}_{birth}$ and ${\bm{\mathcal{M}}}_{birth}$ and append to $X^{(t2t)}_k$;
}
\For{$j$ in $unassigned\_prev\_ind$}{
\% ${\bf{x}}^{(k-1)}_j \in X_{k-1}$ is a decaying target\\
Decrement age, keep the label and append to $X^{(t2t)}_k$, do not update the motion vector; 
}
\% Most mature targets: iterate over $M^{(t2t)}_k$ targets in $X^{(t2t)}_k$\\
\For{$i = 1, 2, \ldots, M^{(t2t)}_k$}{
\If{$a^{(t2t)}_{i} \geq a_T$}{
Append ${{\bf{x}}_{i}^{(t2t)}}^{(k)} \in X_{k-1}$ to $X_k$ \;
}
}
\caption{Target extraction algorithm: The pseudo code explains how the track-to-track (t2t) association and mature targets extraction are performed to obtain $X_k$; $\textrm{RowSum}(\bullet)$, $\textrm{ColSum}(\bullet)$ and $\textrm{HM}(\bullet)$ compute the row-, column-wise summations and Hungarian Assignment, respectively.}
\label{Al:dataAssociation}
\end{algorithm}

{\bf{Note on Kalman update utilisation:}}
The multi-target Kalman update step is only explained to have the paper self-contained. The essential pre-processing step explained in Section \ref{sec:PredImtoEX}, which is utilised to approximate the target states from the non-Gaussian output of ConvLSTM is highly inspired by the unscented Kalman filter's sub-optimal approximation methodology. This results in a closed-form fast MTFT solution, with significantly lower complexity than a similar sequential Monte Carlo (particle filter) approach.  

\subsection{Target state extraction}
After the update step is performed, those targets whose weights are higher than $\omega_T$ are selected as initial updated target RFS $X^{(s)}_k$. Using the target RFS $X_{k-1}$, a track-to-track association is performed over the selected targets. 
The following distance metric is computed between the ${k^{\prime\prime}}^{th}$ target in the previous time step (${\bf{x}}_{k^{\prime\prime}}^{(k-1)} \in X_{k-1}$) and the ${k^{\prime}}^{th}$ (yet unlabelled) target computed at the current time step $\left({{\bf{x}}_{k^\prime}^{(s)}}^{(k)} = \left( {{\bf{m}}_{k^\prime}^{(s)}}^{(k)}, {{\bf{\Sigma}}_{k^\prime}^{(s)}}^{(k)}, {\omega_{k^\prime}^{(s)}}^{(k)} \right) \in X^{(s)}_k\right.$,
\begin{equation}
\left.\delta_{k^{\prime\prime}, k^{\prime}}^{(k)} = - a_{k^{\prime\prime}}^{(k)} \times \textrm{IoU} \left({\bf{x}}_{k^{\prime\prime}}^{(k-1)}, {{\bf{x}}_{k^\prime}^{(s)}}^{(k)} \right)\right),
\end{equation}
where $\textrm{IoU}(\bullet, \bullet) \in 0\leq\mathbb{R}\leq 1$ computes the intersection over union between the two targets. $\delta_{k^{\prime\prime}, k^{\prime}}^{(k)}$ calculates a distance measure between the targets; $a_{k^{\prime\prime}}^{(k)}$ is the age of the target at the previous time step and is multiplied with the $\textrm{IoU}$ to increase the importance of those targets with longer temporal presence. Computing $\delta_{k^{\prime\prime}, k^{\prime}}^{(k)}$ between all the previous and current targets constitutes the matrix ${\bf{\Delta}}_{k, k-1}$, which is given to the Hungarian Assignment algorithm to determine the survival, birth and death of targets, as explained as a pseudo-code in Algorithm~\ref{Al:dataAssociation}: 

\noindent {\bf{Birth of a target:}} A given measurement is considered as a birth if it does not have a corresponding assigned target from the Hungarian matching step (no assignment to any of the targets in the {\bf{MM}} matrix, the output of the {\bf{HM}(.)} function in Algorithm~\ref{Al:dataAssociation}). This case is included in the second loop of the Algorithm~\ref{Al:dataAssociation}. 

\noindent {\bf{Death of a target:}} On the other hand, if there is no association for a target among the measurements and also, its age is lower than $a_T$, the target dies. 

In Algorithm~\ref{Al:dataAssociation}, incrementing (for a survived target) and decrementing (for a decaying target, which does not have any associated target/measurement) the age of the target are performed as follows, 
\begin{align}
\left\{
\begin{array}{l}
a_{k^\prime}^{(k)} = a_{k^\prime}^{(k-1)} + a_{am} \textrm{, if increment is True}\\
a_{k^\prime}^{(k)} = a_{k^\prime}^{(k-1)} - [a_{k^\prime}^{(k-1)} / a_{at}] \textrm{, if decrement is True}
\end{array}
\right.
\label{eq:ageEQ}
\end{align}
\noindent where $[\bullet]$ computes the integer part and $a_{am} \textrm{ and } a_{at} \textrm{ (both in } \mathbb{R^{+}}^{1\times1}$) are the target age amplification and attenuation factors, respectively.
Finally, the most mature targets whose age is higher than a threshold $a_T$ are selected as $X_k$.
$X_k$ is then used to compute the PHD function $v_k$. The PDD map $D_{k, k-1}(x) = v_k - v_{k-1}$ is then calculated and appended to the training batch to train the ConvLSTM network for the next time step.


\subsection{Update time complexity vs. prediction memory complexity}
Assuming no gating step is applied, the complexity of the current update step is $\mathcal{O}(M_{k-1} \times N_{k})$, where $N_{k}$ is the cardinality of $Z_k$ (the number of measurements at $k$). The time complexity of an equivalent particle filter approach can reach $\mathcal{O}(M_{k-1} \times N_{k} \times P)$, where $P$ is the number of particles. 

\noindent Using our implicit {ConvLSTM} paradigm to represent the multi-target state results in a constant memory (spatial) complexity during the prediction step ($\mathcal{O}(1)$). On the other hand, for an explicit representation of the target state used, such as in GM-PHD \cite{Vo:2006}, the memory complexity increases linearly with the number of targets $M_k$ and quadratically with the state space dimensionality $d$ ($\mathcal{O}(d^2\times M_k + d\times M_k)$). For the sequential Monte Carlo (particle filter) approaches, this memory complexity is only linear with the number of targets, such as in PHD-SMC, CPHD-SMC, LMB-SMC and GLMB-SMC \cite{Mahler:2007,Nagappa:2017,Vo:2017,Vo:2014}, as covariance computation is not performed.

\section{Experimental results}
\label{sec:ExpRes}
\subsection{Datasets and algorithm parameters}
We have evaluated our algorithm over the Multiple Object Tracking 2015 (MOT15) and 2016/2017 (MOT16/17) datasets \cite{leal:2015}, which contain 11 and 7 ($\times$ the number of publicly available detectors) video sequences, respectively, captured via both fixed and moving platforms from various crowded scenes for pedestrian tracking\footnote{MOT16 is the MOT17 dataset using only the Deformable Parts Model (DPM) detector}. The pedestrian detection is performed by: Aggregated Channel Features (ACF) \cite{Dollar:2014} for MOT15; DPM \cite{Felzenszwalb:2010}, Faster Region-based CNN (FRCNN) \cite{Ren:2015} and Scale Dependent Pooling (SDP) \cite{Yang:2016} for MOT17.
It should be pointed out that we have chosen to use the MOT15 dataset to evaluate the multi-target filtering performance, because of the high intensity of clutter generated by ACF over this dataset. 
We have also used the Performance Evaluation of Tracking and Surveillance 2009 S2 (PETS09) sequences (\url{http://www.cvg.reading.ac.uk/PETS2009/}) with an ACF pedestrian detector. The PNNL Parking Lot 1~\cite{Shu:2012}, the only sequence of PNNL dataset with publicly available detections, is also utilised. 
The proposed MTFT algorithm is implemented, end-to-end, in Python 2.7. Keras with a Tensorflow backend is used for the ConvLSTM implementation, over an NVIDIA GeForce GTX 1080 Ti GPU, where the average frame per second of the proposed algorithm is $\approx 14$ fps. 
Our following results are obtained using $N = 24$, ReLU activation function, one block of sixteen $3 \times 3$ convolutional filters, 20 training epochs, $B = 5$ pixels, $p_{D, k} = 0.9$ for all detection algorithms, $\omega_T = 0.5$, $a_T = a_{birth} = 5$, $a_{at} = 2$, ${\bm{\mathcal{M}}}_{birth} = [0, 0]$, $a_{am} = 1$ and $\Sigma_{birth} = 20*{\bf{I}}_4$, where ${\bf{I}}_4$ is a $4\times4$ identity matrix. During the training of the ConvLSTM network we used the ADAM optimiser, with default parameters (learning rate, $\beta_1$, and $\beta_2$ as
0.001, 0.9, and 0.99, respectively). 
We use the following strategies to overcome lack of data and under-/over-fitting during the ConvLSTM network training: (1) Feature engineering: instead of feeding raw PHD maps we defined the PDD maps, resulting significant improvement on the learning convergence; (2) Avoiding too deep architecture; (3) Use of weight  regularisation (activity, kernel and bias); (4) Early stopping approach: reducing the maximum number of epochs; (5) KL divergence loss function, which outperformed other objective functions, especially when the network's outputs were sparse. 
Our Supplementary Material provides some video samples showing the tracking results and explanations on how the evaluation metrics are calculated.

\subsection{Filtering and tracking performance}
\label{sec:FiltTrack}
As OSPA assumes point targets, here we used the centre of the bounding boxes to represent each target (in the next section, however, we evaluate the tracking performance using bounding box representation for targets). 

Table~\ref{tab:table1} and \ref{tab:table2} show comparative performance of several MTFT algorithms over the MOT15 and MOT16/17 datasets, respectively. As OSPA is computed at each time step, the results in these tables are the average over all time steps for all video sequences (it should be mentioned that Loc OSPA can be easily computed by subtracting the overall and Card errors).
Considering the PHD-based algorithms (PHD-EKF, PHD-SMC and PHD-UKF) as baseline, our proposed ConvLSTM MTFT algorithm shows significantly better performance over both datasets. Particularly there is remarkable reduction in the cardinality error.
We have also compared our method with four tracking algorithms: SORT \cite{Bewley:2016}, which is one of the fastest online tracking approaches reported over MOT15, DeepSORT \cite{wojke2017simple} an extension of SORT with deep associations, Re$^3$ \cite{Gordon:2018}, a deep RNN-based multi-target tracker, and RNN\_LSTM algorithm \cite{Milan:2017}, one of the pioneering algorithms using RNN for multi-target tracking. Our overall average OSPA error is $\approx 1.12$, $2.40$, $5.65$ and $5.71$ lower than RNN\_LSTM, SORT, DeepSORT and Re$^3$, respectively. 
For the case of MOT16/17, the overall average error is higher than MOT15 results for all algorithms (Table \ref{tab:table2}). The reason is that compared to MOT15, there are significantly higher number of annotated objects in this dataset. However, similar to MOT15, our algorithm outperforms the other methods in terms of overall OSPA, with $\approx 8.5$ Loc and $\approx 53.5$ Card errors.
\begin{table}[!t]
\footnotesize{
\begin{center}
\begin{tabular}{l l l} \hline
Methods & OSPA Card & OSPA \\
\hline
PHD-EKF & $29.3 \pm 10.2$ & $46.9 \pm 13.5$ \\
PHD-SMC & $25.7 \pm 9.0$ & $52.9 \pm 13.0$ \\
PHD-UKF & $28.7 \pm 10.0$ & $46.5 \pm 13.2$ \\
\hdashline
CPHD-EKF & $23.7 \pm 9.2$ & $44.7 \pm 12.8$ \\
CPHD-SMC & $51.1 \pm 15.5$ & $75.2 \pm 12.9$ \\
CPHD-UKF & $23.6 \pm 8.8$ & $44.7 \pm 12.0$ \\
\hdashline
LMB-EKF & $23.9 \pm 8.3$ & $49.5 \pm 14.2$ \\
LMB-SMC & $62.2 \pm 29.5$ & $71.5 \pm 23.8$ \\
LMB-UKF & $24.4 \pm 7.4$ & $48.8 \pm 12.6$ \\
\hdashline
GLMB-EKF & $93.4 \pm 8.5$ & $95.0 \pm 7.6$ \\
GLMB-SMC & $90.4 \pm 10.8$ & $92.2 \pm 9.00$ \\
GLMB-UKF & $48.1 \pm 19.4$ & $57.4 \pm 19.3$ \\
\hdashline
SORT & $29.4 \pm 18.8$ & $41.5 \pm 16.7$ \\
DeepSORT & $31.4 \pm 21.1$ & $46.1 \pm 18.4$\\
Re$^3$ & $34.4 \pm 21.2$ & $46.1 \pm 18.7$ \\
RNN$\_$LSTM & $23.7 \pm 18.4$ & $42.8 \pm 15.7$ \\
\hdashline
\textbf{ConvLSTM} & $\textbf{19.2} \pm \textbf{5.1}$ & $\textbf{40.4} \pm \textbf{11.4}$ \\ \hline
\end{tabular}
\end{center}
}
\caption{OSPA error $\pm$ standard deviation on MOT15: Comparison against PHD \cite{Mahler:2007,Vo:2006}, CPHD \cite{Nagappa:2017,Mahler:2007}, LMB \cite{Reuter:2014} and GLMB \cite{Vo:2014,Vo:2017}, with EKF, SMC, and UKF prediction and update steps, and other multi-target tracking algorithms.}
\label{tab:table1}
\end{table}

\begin{table}[!t]
\footnotesize{
\begin{center}
\begin{tabular}{l l l} \hline
Methods & OSPA Card & OSPA \\
\hline
PHD-EKF & $60.5 \pm 15.6$ & $66.9 \pm 13.3$ \\
PHD-SMC & $49.3 \pm 19.5$ & $67.2 \pm 13.1$ \\
PHD-UKF & $60.1 \pm 15.7$ & $66.5 \pm 13.4$ \\
\hdashline
CPHD-EKF & $56.8 \pm 18.9$ & $65.4 \pm 14.9$ \\
CPHD-SMC & $46.5 \pm 19.2$ & $78.7 \pm 11.3$ \\
CPHD-UKF & $57.00 \pm 18.9$ & $65.2 \pm 14.9$ \\
\hdashline
LMB-EKF & $57.8 \pm 19.5$ & $70.4 \pm 14.2$ \\
LMB-SMC & $90.0 \pm 19.0$ & $92.0 \pm 15.0$ \\
LMB-UKF & $77.9 \pm 12.3$ & $82.8 \pm 9.3$ \\
\hdashline
GLMB-EKF & $94.8 \pm 4.3$ & $95.5 \pm 3.7$ \\
GLMB-SMC & $93.9 \pm 6.1$ & $94.8 \pm 5.0$ \\
GLMB-UKF & $82.7 \pm 14.6$ & $85.0 \pm 12.8$ \\
\hdashline
SORT & $61.7 \pm 10.7$ & $70.4 \pm 8.6$ \\
DeepSORT & $63.4 \pm 10.0$ & $71.3 \pm 8.3$\\
Re$^3$ & $62.1 \pm 10.5$ & $71.4 \pm 8.3$ \\
RNN$\_$LSTM & $60.4 \pm 10.3$ & $70.3 \pm 7.9$ \\
\hdashline
\textbf{ConvLSTM} & $\textbf{53.8} \pm \textbf{18.3}$ & $\textbf{62.3} \pm \textbf{14.4}$ \\\hline
\end{tabular}
\end{center}
}
\caption{OSPA error $\pm$ standard deviation on the MOT16/17 dataset.}
\label{tab:table2}
\end{table}

\begin{table}[!h]
\centering
\begin{tabular}{lllll}
\hline 
Algorithm & Rcll & Prcn & MOTA & MOTAL \\
\hline 

{SORT-DPM} & {37.0} & {76.6} & {24.7} & {25.7} \\
{SORT-FRCNN} & {50.5} & {97.3} & {48.5} & {49.2} \\
{SORT-SDP} & {63.0} & {\textbf{98.3}} & {61.0} & {61.8} \\\hdashline

{DeepSORT-DPM} & {32.4} & {91.3} & {28.2} & {29.3} \\
{DeepSORT-FRCNN} & {51.7} & {95.7} & {48.6}  & {49.4} \\
{DeepSORT-SDP} & {64.7} & {97.3} & {61.6} & {62.9} \\\hdashline

{Re$^3$-DPM} & {37.0} & {76.2} & {24.3} & {25.4} \\
{Re$^3$-FRCNN} & {50.6} & {97.1} & {48.4} & {49.0} \\
{Re$^3$-SDP} & {62.9} & {98.0} & {60.7} & {61.6} \\
\hdashline

{RNN{\_}LSTM-DPM} & {32.8} & {83.6} & {25.6} & {26.4} \\ 
{RNN{\_}LSTM-FRCNN} & {44.7} & {88.9} & {38.4} & {39.1} \\
{RNN{\_}LSTM-SDP} & {49.7} & {87.7} & {41.5} & {42.7} \\
\hdashline
                                 
{ConvLSTM-DPM} & {38.9} & {69.9} & {20.3} & {22.1} \\
{ConvLSTM-FRCNN} & {53.3} & {92.6} & {48.1} & {49.0} \\
{\textbf{ConvLSTM-SDP}} & {\textbf{67.1}} & {94.9} & {\textbf{62.0}} & {\textbf{63.5}} \\
\hline                                 
\label{tab:Mot17}
\end{tabular}
\caption{Multi-target tracking performance over MOT16/17.}
\label{tab:Mot17}
\end{table}

\begin{table}[!h]
\centering
\begin{tabular}{lllll}
\hline 
Algorithm & Rcll & Prcn & MOTA & MOTAL \\
\hline {SORT} & {71.4} & {\textbf{98.5}} & {69.0} & {70.3} \\
{DeepSORT} & {74.3} & {97.5} & {70.5} & {\textbf{72.1}} \\
{Re$^3$} & {70.9} & {97.8} & {68.0} & {69.3} \\
\hdashline
{\bf{ConvLSTM}} & {\textbf{77.7}} & {93.3} & {\textbf{71.0}} & {\textbf{72.1}} \\
\hline
\end{tabular}
\caption{Multi-target tracking performance over PNNL Parking Lot dataset using~\cite{Shu:2012} detector (RNN{\_}LSTM is excluded as its pre-trained models are not provided for this dataset).}
\label{tab:PNNL}
\end{table}

\begin{table}[!h]
\centering
\begin{tabular}{lllll}
\hline 
Algorithm & Rcll & Prcn & MOTA & MOTAL \\
\hline {SORT} & {75.0} & {\textbf{87.4}} & {61.9} & {64.1} \\
{DeepSORT} & {88.0} & {83.0} & {65.2} & {62.9} \\
{Re$^3$} & {73.1} & {83.6} & {56.5} & {58.8} \\
{RNN{\_}LSTM} & {\textbf{91.1}} & {68.1} & {43.9} & {48.3} \\
\hdashline
{\textbf{ConvLSTM}} & {80.9} & {86.8} & {\textbf{66.9}} & {\textbf{68.5}} \\
\hline
\end{tabular}
\caption{Multi-target tracking performance over the PETS09 dataset using ACF \cite{Dollar:2014} as the detector.}
\label{tab:PETS09}
\end{table}

Three datasets are used to quantitatively evaluate the tracking performance of the proposed algorithm: MOT16/17, PNNL Parking Lot and PETS09. Unlike the OSPA results, for which point targets were assumed, here each target is represented as a bounding box. 
The multi-target tracking results for MOT16/17, PNNL Parking Lot and PETS09 are shown in Tables~\ref{tab:Mot17}, \ref{tab:PNNL}, \ref{tab:PETS09}, respectively. For MOT16/17, all three publicly available detections are used. ConvLSTM's performance is at the highest when the SDP detector is used, with $\approx 62\%$ MOTA, $\approx 67.1\%$ recall and $\approx 62\%$ MOTAL. Also, ConvLSTM generates $\approx 71.0\%$ and $\approx 66.9\%$ MOTA, over the Parking Lot and PETS09, when the detection method in \cite{Shu:2012} and ACF are used, respectively. For the Parking Lot sequence, ground truth is only available for the first 748 frames, and hence we have evaluated the algorithms over these frames only.
Our ConvLSTM approach generates lower miss rate (false negatives), resulting in higher recall, as it can be seen in Tables~\ref{tab:Mot17} and \ref{tab:PNNL}. We also computed the standard deviation of the MOTA metric calculated over 10 runs of the algorithm. For the MOT16/17 results shown in Table \ref{tab:Mot17}, the standard deviation for DPM, FRCNN and SDP detector are 0.09\%, 0.05\% and 0.06\%, while for PNNL (Table \ref{tab:PNNL}) and PETS09 (Table \ref{tab:PETS09}) are 0.40\% and 0.76\%, respectively. These very low standard deviations show very high consistency of the proposed ConvLSTM algorithm.

\subsection{Further standard deviation analysis} 
\begin{figure}[!t]
    \centering
    \begin{subfigure}[]{0.5\textwidth}
        \includegraphics[width=\textwidth]{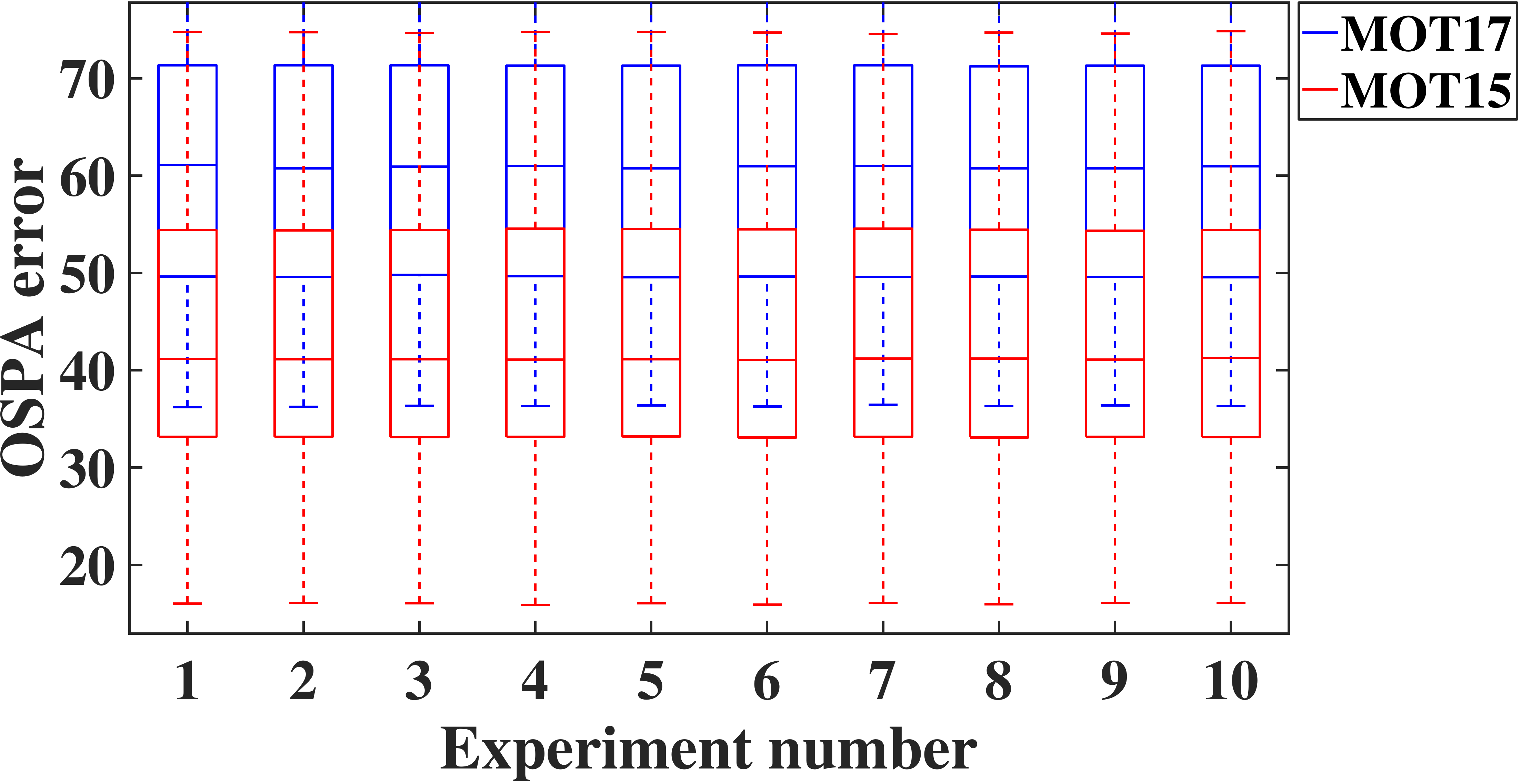}
        \caption{}
        \label{fig:OSPABox}
    \end{subfigure}\\
    \begin{subfigure}[]{0.5\textwidth}
        \includegraphics[width=\textwidth]{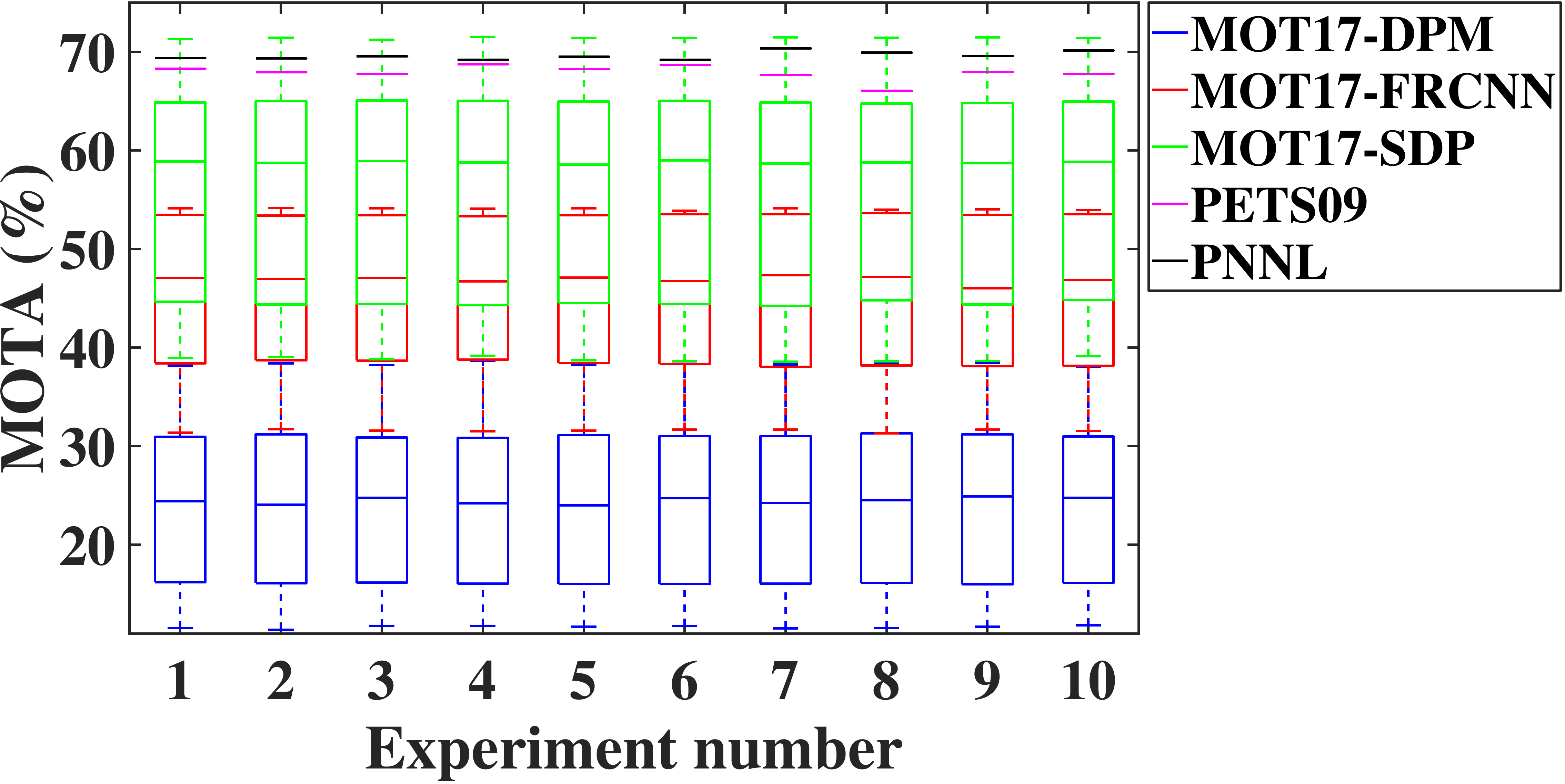}
        \caption{}
        \label{fig:MOTABox}
    \end{subfigure}
    \caption{ANOVA test over: (a) OSPA for MOT15 and MOT16/17; and (b) MOTA for MOT16/17, PNNL and PETS09 datasets. The box-plots are calculated for several experiment runs, showing very low sensitivity to the random parameters in the algorithm, as the average OSPA and MOTA are very consistent for all experiments.}
    \label{fig:ANOVA}
\end{figure}

In order to evaluate the robustness of the {ConvLSTM} algorithm, we perform the one-way analysis of variance (ANOVA). ANOVA determines the statistical significance between the means of independent groups. 
For this, we conducted an experiment in which we evaluate the repeatability of the generated results and test how invariant the outputs are against the algorithm's randomness. Such randomness can be originated from either the neural network training step (for example, its random weights and biases initialisation or batch generation) or the parameters used throughout the filtering stage (such as clutter density estimation). As the proposed MTFT algorithm is trained on the fly, this consistency evaluation shows how repeatable the generated results are.
We therefore repeat the whole end-to-end MTFT solution 10 times, for each dataset.
The resulting OSPA and MOTA are then used to compute the box-plots shown in Fig.~\ref{fig:ANOVA}-a and -b, respectively.
As seen in these figures, the results show very low sensitivity to the random parameters in the algorithm, as the average
OSPA, together with its percentiles are similar among the different
experiments for both MOT15 and MOT17 dataset. 
Correspondingly, the average MOTA is also computed among the different videos. For each detector in MOT17, samples in the PNNL and PETS09 datasets, similar consistency is observed from Fig.~\ref{fig:ANOVA}-b.

\begin{figure}[!t]
    \centering
    \begin{subfigure}[]{0.5\textwidth}
        \includegraphics[width=\textwidth]{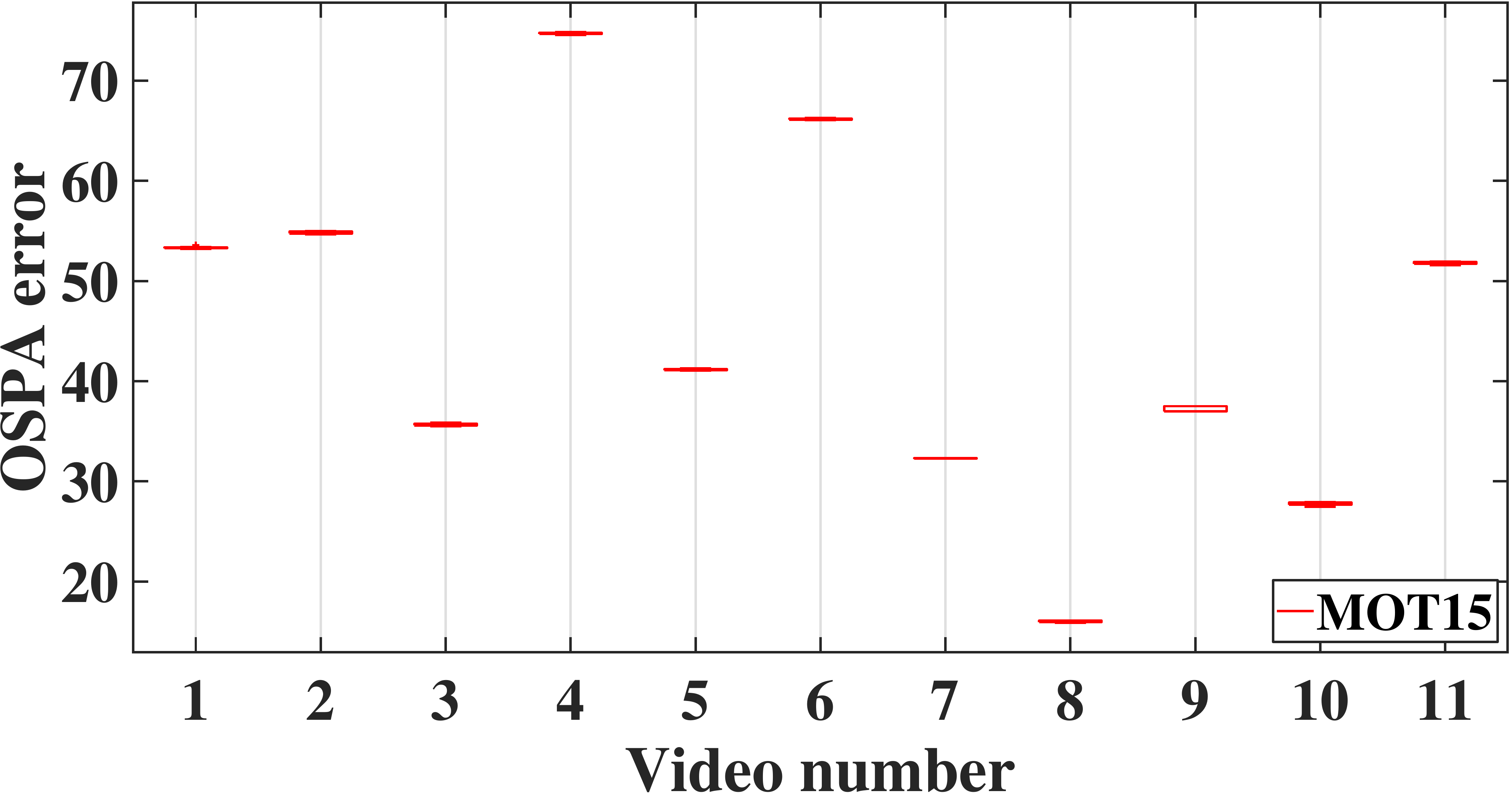}
        \caption{}
        \label{fig:OSPAMot15}
    \end{subfigure}\\
    \begin{subfigure}[]{0.5\textwidth}
        \includegraphics[width=\textwidth]{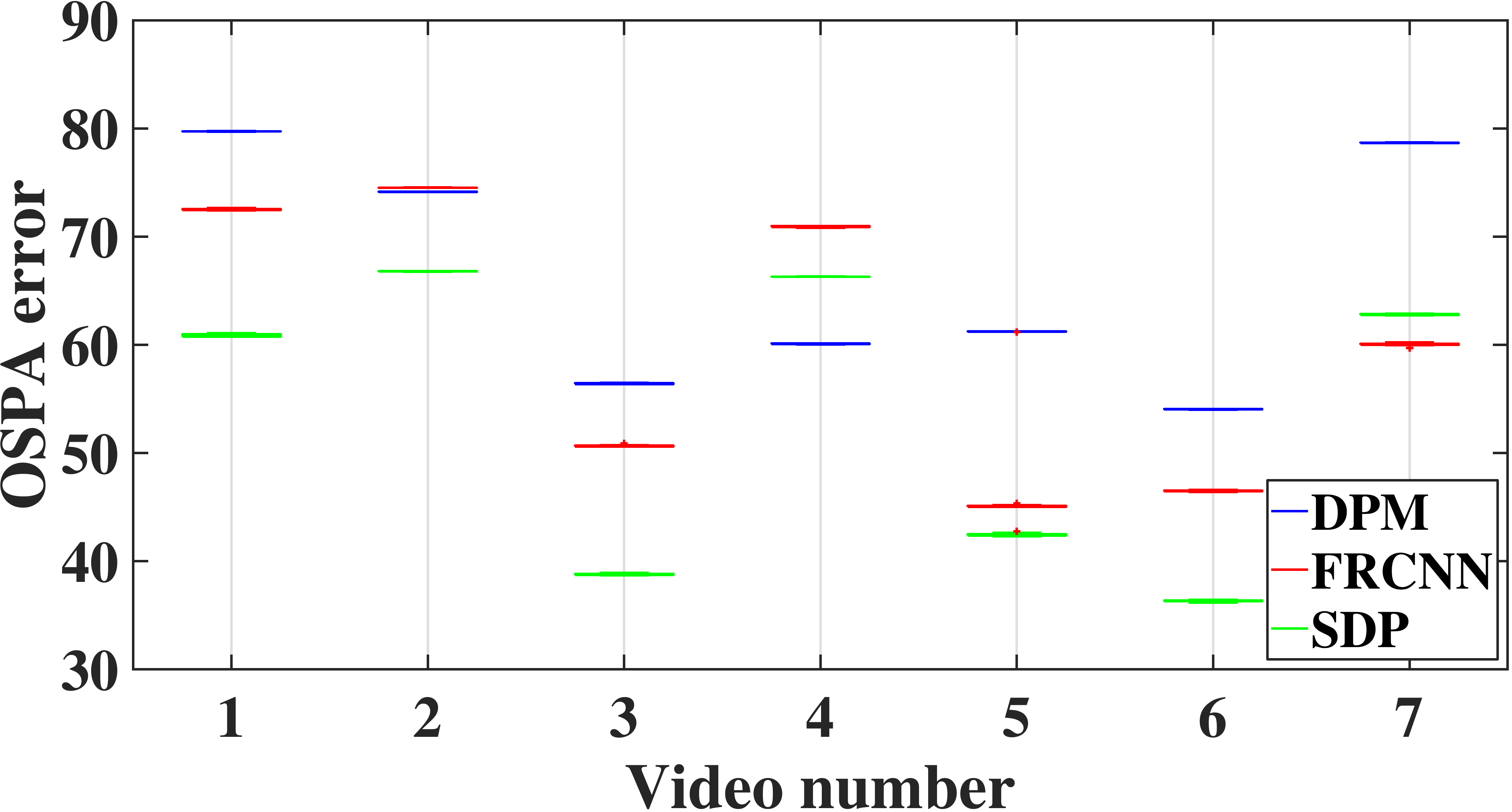}
        \caption{}
        \label{fig:OSPAMot17}
    \end{subfigure}
    \caption{ANOVA test over OSPA for every video sequence of: (a) MOT15 and (b) MOT16/17, after 10 runs. The very narrow width of the box-plots shows very high repeatability for each video sequence.}
    \label{fig:VidNumVsOSPA}
\end{figure}

The experiment in Fig.~\ref{fig:ANOVA}-a is also expanded for each video frame within the MOT15 and 16/17 datasets. The resulting OSPA metric is then averaged and the standard deviation is calculated. The result is shown in Fig.~\ref{fig:VidNumVsOSPA}-a and -b, for MOT15 and MOT16/17, respectively, for each of their video sequences. For each individual video sequence, the standard deviation is very low, resulting in box-blots with very narrow width (It should be mentioned that similar experiment can not be performed for MOTA, as unlike OSPA, MOTA is not averaged over frames).

\subsection{Study on hyper-parameter sensitivity}
\label{sec:Sen}
\begin{figure}[!t]
    \centering
    \begin{subfigure}[]{0.4\textwidth}
        \includegraphics[width=\textwidth]{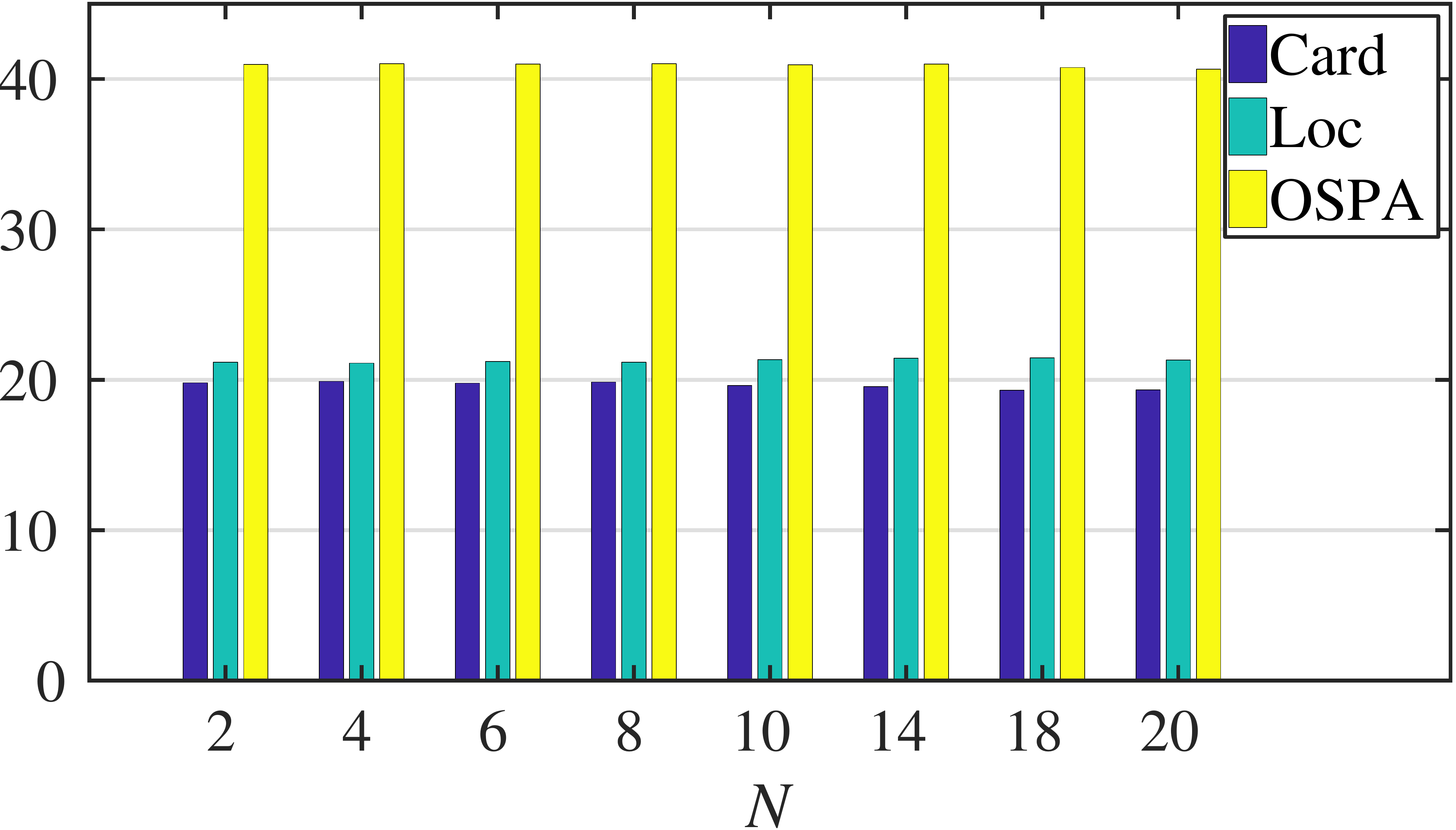}
        \caption{}
        \label{fig:batchSize}
    \end{subfigure}\\
    \begin{subfigure}[]{0.4\textwidth}
        \includegraphics[width=\textwidth]{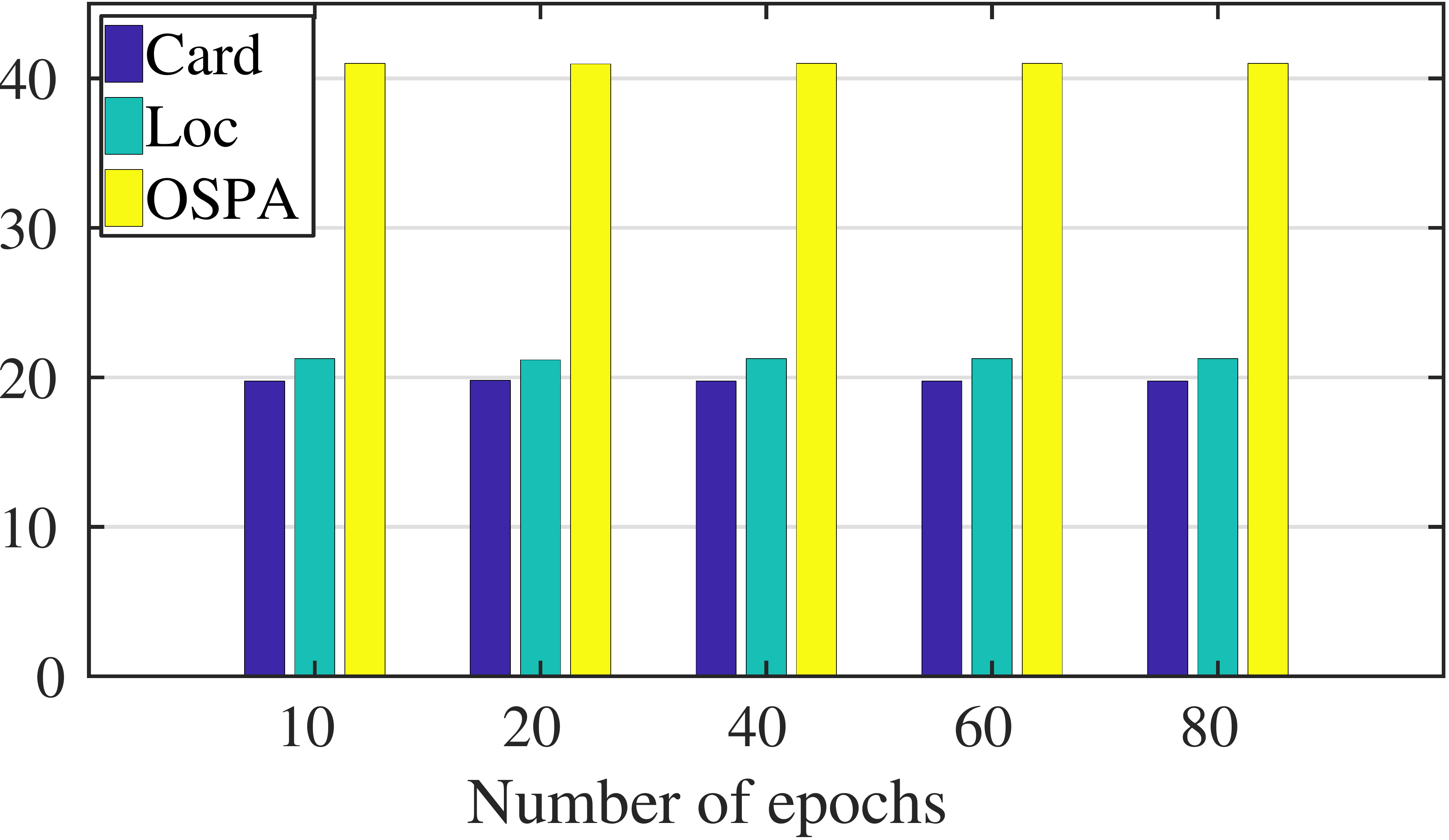}
        \caption{}
        \label{fig:NEpochs}
    \end{subfigure}
    \caption{The OSPA error when: (a) batch size and (b) number of training epochs are varied. As our algorithm shows very low sensitivity to these parameters, lower values can be used to increase the speed.}
    \label{fig:NNSensitivity}
\end{figure}

In this section, we evaluate the sensitivity of the proposed algorithm to various hyper-parameters and functions.
In the first set of results of this section, the average Loc, Card and overall OSPA errors over all MOT15 videos are calculated, when three hyper-parameters are changed: 1) The PDD batch size ($N$), 2) number of training epochs, and 3) target age attenuation factor $a_{at}$. The results of the first two are shown in Fig.~\ref{fig:NNSensitivity}-a and -b, respectively. The standard deviation of the overall OSPA error when $N$ and number of epochs are increased are $\approx 0.12$ and $0.01$, respectively, which shows the algorithm performance has very low sensitivity to these two parameters. Such low sensitivity to $N$ can be due to two reasons: 1) The use of the LSTM architecture, which is capable of memorising long-term dependencies and 2) representing the latent space via the PHD functions, which encapsulate the temporal information over the hypothetical target state. This low sensitivity to $N$ and number of epochs, however, is an important fact as it indicates smaller values can be chosen for these two parameters, without reduction in the accuracy. This can significantly increase the computation time.

\begin{figure}[!t]
\centering
\includegraphics[width=0.5\textwidth]{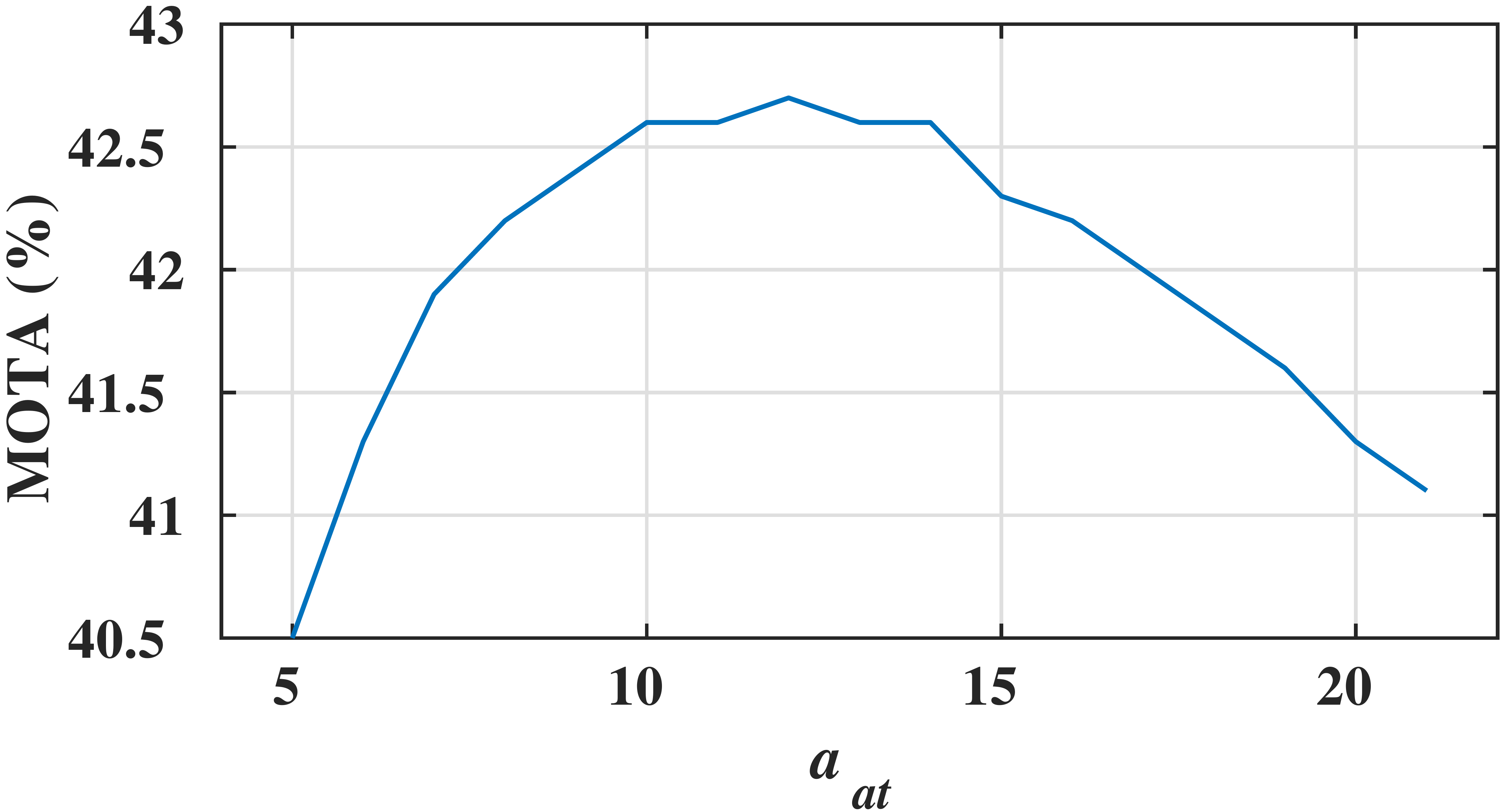}
\caption{MOTA of the ConvLSTM MTFT algorithm when $a_{at}$ is varied.}
\label{fig:attResMOTA}
\end{figure}

Figure \ref{fig:attResMOTA} illustrates the MOTA results when the target age attenuation factor ($a_{at}$) is varied. The MOTA reported for this figure is computed by averaging all of the MOTA results for all MOT16/17 sequences over its three different detectors. When $a_{at}$ is increased, the targets' age decrement rate is reduced. This may create false positive targets and consequently, lower MOTA is generated. On the other hand, for lower $a_{at}$, the targets' age is more rapidly reduced. This can create false negative (high miss rate) and hence lower MOTA as well. Both of these cases are observed from Fig.~\ref{fig:attResMOTA}. As a trade-off between these two scenarios, $a_{at} \approx 13$ generates the highest overall MOTA error.

\begin{figure}[!t]
    \centering
    \begin{subfigure}[]{0.5\textwidth}
        \includegraphics[width=\textwidth]{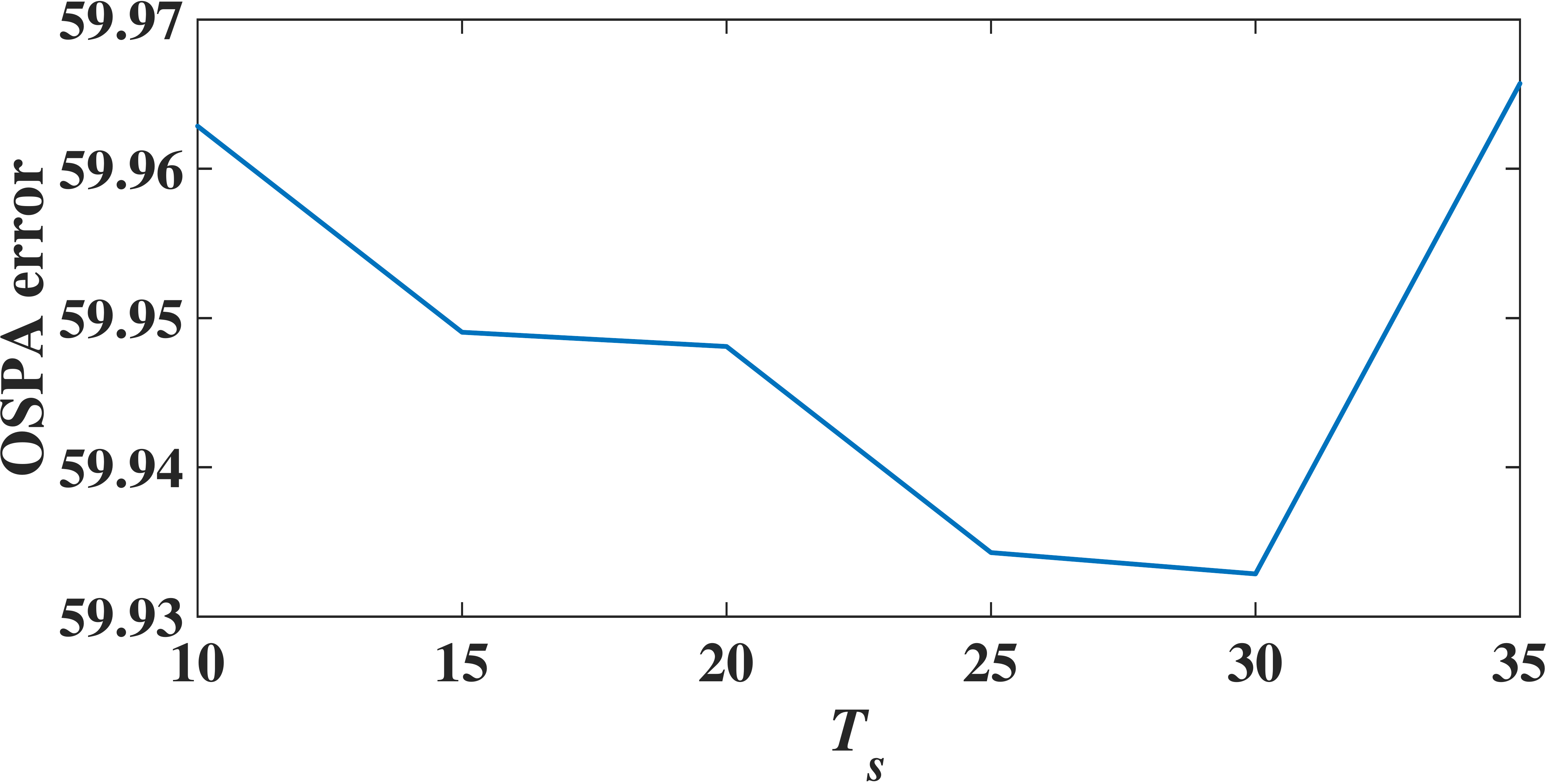}
        \caption{}
        \label{fig:OSPAGran}
    \end{subfigure}\\
    \begin{subfigure}[]{0.5\textwidth}
        \includegraphics[width=\textwidth]{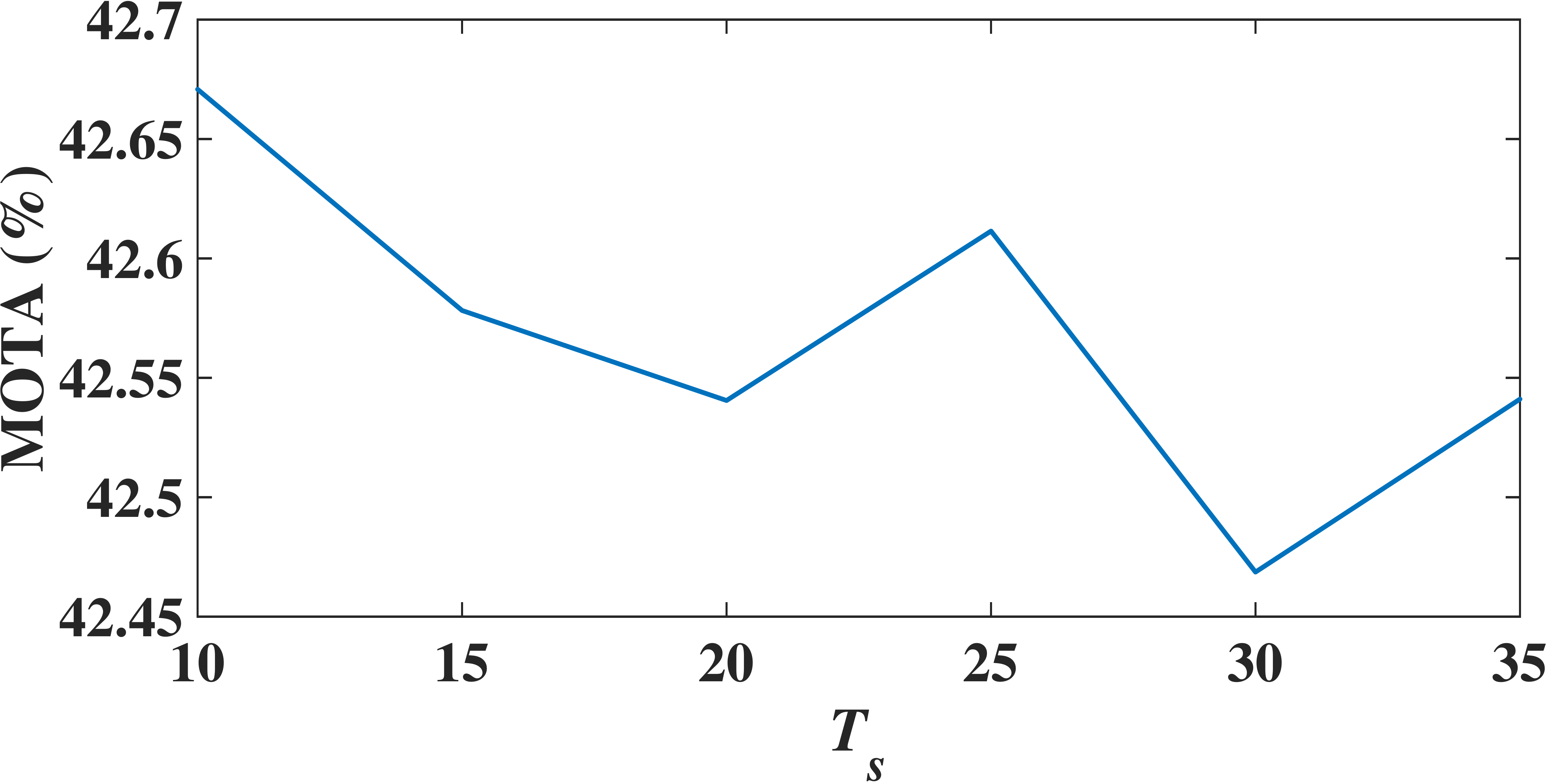}
        \caption{}
        \label{fig:MOTAGran}
    \end{subfigure}
    \caption{The filtering and tracking performance evaluation over the MOT16/17 dataset against various sampling periods $T_s$: (a) OSPA and (b) MOTA results.}
    \label{fig:granularityAnalysis}
\end{figure}

One of the key hyper-parameters of the proposed {ConvLSTM} MTFT algorithm is the sampling period $T_s \triangleq \frac{1}{f_s}$, which, by definition, is the inverse of sampling frequency $f_s$ (the unit of $f_s$ is: $\frac{\textrm{sample}}{\textrm{state unit}}$). $T_s$ is used to discritise the state space, in order to create the PHD map. A low sampling period results in a finer state increments, the likelihood of target merge is reduced and hence can be more representative of the multi-target state. However, a very low sampling period (high sampling frequency) increases the computational complexity, as the ConvLSTM network is trained over a larger map. Also, the state space will be more sparse resulting in a flatter loss function, which can fail the ConvLSTM optimiser to find a local minimum. As is shown in Fig.~\ref{fig:granularityAnalysis}, this has different effects over filtering and tracking performance. Figure~\ref{fig:granularityAnalysis}-a shows that for sampling rates $< 20$, the OSPA error increases. This is due to the fact that for low $T_s$ (a denser PHD map), the probability of obtaining wrong number of targets increases as the false positive rate grows. This increases the CARD error of the OSPA error. However, as can be seen Fig.~\ref{fig:granularityAnalysis}-b, MOTA increases for this range of $T_s$, which can be because of decrease in obtaining false negative rate.

On the other hand, a very high sampling period can merge several targets, resulting in a {\emph{blurry}} PHD map. While this is less computationally problematic for the predictor (ConvLSTM neural network), it significantly deteriorates the performance of the data association step. This has a negative effect over both the filtering and tracking performance as can be seen in the increase of OSPA in Fig.~\ref{fig:granularityAnalysis}-a and decrease of MOTA in Fig.~\ref{fig:granularityAnalysis}-b, for $T_s > 30$. It should, however, be mentioned that the variation of OSPA and MOTA over various $T_s$ is very low (the vertical axis of both Fig.~\ref{fig:granularityAnalysis}-a and -b).

Finally, we also evaluate the effect of varying the loss function from KL to Jensen-Shannon divergence (JSD). Same as KL divergence, JSD is also a technique of measuring the similarity between two probability distributions; however, unlike the KL divergence, JSD is a metric and symmetrical, in the sense that it is invariant to the order of input arguments. The filtering and tracking performance of using KL divergence and JSD is illustrated in Table~\ref{tab:JSD}, showing extremely similar performance. KL divergence is used as the neural network's loss (objective) function, consistently, applied over all of the samples during the training phase. The order of KL divergence input arguments remained the same throughout the training process. During the optimisation, we are only interested in the relative amplitude of the loss function between the epochs in order to find the local/global minima. Therefore, due to the close similarity between the prediction and ground truth data frames (input arguments for both the KL divergence and JSD), their computed minima stay very close and as a result the performance of JSD becomes very similar to KL (Table~\ref{tab:JSD}).

\begin{table}[!h]
\centering
\begin{tabular}{lll}
\hline 
Dataset & KL & JSD \\\hline 
\multirow{5}{*}{{MOT16/17 DPM}} & Rcll: 38.5 & Rcll: 38.5\\
 & Prcn: 71.1 & Prcn: 71.1\\
 & MOTA: 21.6 & MOTA: 21.5\\
 & MOTAL: 22.9 & MOTAL: 22.8\\
 & OSPA: 66.34 & OSPA: 66.33\\
\hline 
\multirow{5}{*}{{MOT16/17 FRCNN}} & Rcll: 51.9 & Rcll: 51.8\\
 & Prcn: 92.0 & Prcn: 92.2\\
 & MOTA: 46.5 & MOTA: 46.6\\
 & MOTAL: 47.4 & MOTAL: 47.4\\
 & OSPA: 60.02 & OSPA: 60.0\\
\hline 
\multirow{5}{*}{{MOT16/17 SDP}} & Rcll: 65.0 & Rcll: 65.0\\
 & Prcn: 94.3 & Prcn: 94.1\\
 & MOTA: 59.8 & MOTA: 59.6\\
 & MOTAL: 61.1 & MOTAL: 60.8\\
 & OSPA: 53.46 & OSPA: 53.5\\
\hline 
\end{tabular}
\caption{Comparing the filtering and tracking performance when JSD is used as the loss function (third column) instead of KL divergence (second column). In order to verify the results are not biased by the detector, all of the MOT16/17 detectors are evaluated at each row (further mathematical analysis provided in out Supplementary Material).}
\label{tab:JSD}
\end{table}

\section{Conclusions}
\label{sec:conc}
This paper detailed a spatio-temporal data prediction approach applicable for MTFT problems. The prediction is simultaneously performed for all of the targets, over an implicit continuous hypothetical target space, via ConvLSTM neural network. The proposed approach not only significantly improves the baseline RFS filters, but also shows substantial potential when compared with other state-of-the-art tracking algorithms. 

Our approach is able to learn complex target state changes by means of the use of a ConvLSTM predictor network over PHD maps, which learns the motion model over time. This capability explains the high performance achieved in the experiments. As a result of this learning step, the algorithm outperforms those approaches that are based on fixed motion models. Also, the use of PHD maps, LSTM architecture, target tuple definition in Section \ref{sec:PDD}, and the data association step enable the algorithm to memorise the state of the targets, reducing the probability of ID switches and sensitivity to occlusions.

Our algorithm, which makes an important step towards an end-to-end learning model for online MTFT over realistic scenarios, can be enhanced in several aspects. An immediate improvement can be to include the update step within the {ConvLSTM} framework. Future work will be focused on the inclusion of the update step within the ConvLSTM framework. Furthermore, as an alternative for the ConvLSTM network (which is a discriminative algorithm approximating the posterior densities), generative neural networks can be utilised to directly estimate the joint PDF, eliminating the need for a separate covariance estimation at the prediction step.

\begin{IEEEbiography}[{\includegraphics[width=1in,height=1.25in,clip,keepaspectratio]{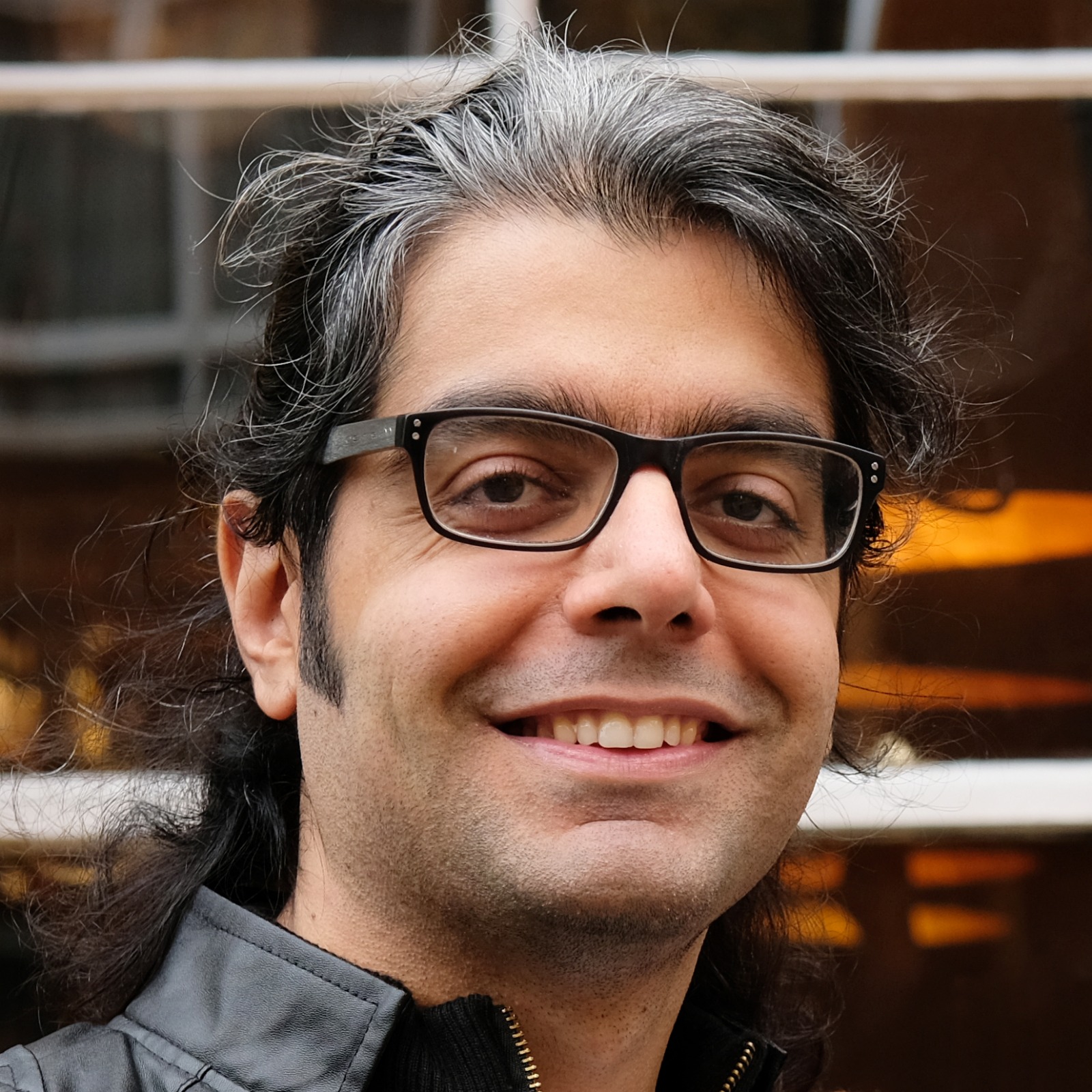}}]%
{Mehryar Emambakhsh}
received the PhD degree in electronic and electrical engineering from the University of Bath in 2015, researching the potential of the {3D} shape of the nose for biometric authentication. He was a Post-doctoral Research Associate in big data science at Aston University and in radar and video data fusion for autonomous vehicles at Heriot-Watt University from 2014 to 2017. He was a Research Scientist at Cortexica Vision Systems Ltd from 2018 to 2019, developing multi-target filtering and tracking algorithms. He is currently a Vice President and Senior Research Scientist at Mesirow Financial. His research interest is in designing algorithms to learn from sequential data.
\end{IEEEbiography}

\begin{IEEEbiography}
[{\includegraphics[width=1in,height=1.25in,clip,keepaspectratio]{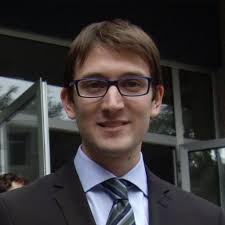}}]%
{Alessandro Bay}
received the B.S. degree in Mathematics for Engineering in 2011, the M.S. degree in Mathematics Engineering in 2013, and the Ph.D. degree in Electronics Engineering, in 2017, all from Politecnico di Torino, Turin, Italy. He is currently a Research Scientist, working at Cortexica Vision Systems Ltd, London, UK. His main research interests include computer vision, deep learning, and mathematical modelling.
\end{IEEEbiography}

\begin{IEEEbiography}
[{\includegraphics[width=1in,height=1.25in,clip,keepaspectratio]{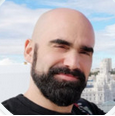}}]%
{Eduard Vazquez} received his PhD degree in computer vision from Universitat Aut\`{o}noma de Barcelona. He has been working in this field for the last 15 years, having 10 years experience on commercialising a variety of methods for medical imaging, retail, manufacturing, surveillance, health and safety and real-time video analytics. His main focus has been to understand the gap between theory and the real world, in order to help delivering products that can be effectively adopted. He is currently the Research Technical Manager at Anyvision, where he works on face recognition for multiple use cases and industries and in-store analytics for retail, from cloud to edge.
\end{IEEEbiography}

%
\appendices

%
%
%

%
%

\markboth{IEEE Transactions on Signal Processing}%
{Emambakhsh \MakeLowercase{\textit{et al.}}: Convolutional Recurrent Predictor: Implicit Representation for Multi-target Filtering and Tracking}
%



\maketitle

%


%
\IEEEpeerreviewmaketitle
\section{Loss function definition}
During training, a Kullback-Leibler (KL) divergence loss is minimised, defined as follows,
\begin{align}\label{eq:KLDiv}
&\mathcal{D}_{KL}\left(D_{k-i, k-i-1} || D^{pred, (t)}_{k-i, k-i-1} \right) =\\
&\int \overline{D}_{k-i, k-i-1}(x) \log\left( \frac{\overline{D}_{k-i, k-i-1}(x)}{\overline{D_{k-i, k-i-1}^{pred, (t)}}(x)} \right)dx, \nonumber
\end{align}
in which $\overline{D}_{k-i, k-i-1}(x)$ and $\overline{D_{k-i, k-i-1}^{(pred), (t)}}(x)$ are the normalised (integrated to one) versions of ${D}_{k-i, k-i-1}(x)$ and ${D_{k-i, k-i-1}^{(pred), (t)}}(x)$. 
${D}_{k-i, k-i-1}(x)$ is computed using the PHD functions at the $(k-i)^{th}$ and $(k-i-1)^{th}$ time steps and ${D_{k-i, k-i-1}^{(pred), (t)}}(x)$ is the predicted target states at the $t^{th}$ epoch. 

\section{KL vs. Jensen-Shannon divergence}
In this section, we compare the differentiation of KL with the Jensen-Shannon divergence (JSD), when used as the neural network objective function. We assume $q = q(x)$ is the variable prediction signal generated by the network and $p = p(x)$ is the constant ground truth data. $x$ represents the corresponding parameter in the neural network (for example, weights or biases) and for simplicity, we assume it is scalar. The differentiation of the KL divergence with respect to $q$ can be computed as follows,
\begin{align*}
&\mathcal{D}_{KL}\left(p || q \right) = \int p \log(\frac{p}{q}) dx,\\
&\frac{d\mathcal{D}_{KL}}{dq} = \int{\frac{d}{dq}\left(p\log(\frac{p}{q})\right)dx} = \int{-\frac{p}{q}}dx,
\end{align*}
\noindent where $\mathcal{D}_{KL}\left(p || q \right)$ is the KL divergence loss function. On the other hand, the differentiation of a JSD objective function with respect to $q$ will be (for which we have removed the 0.5 scale factor, as it will not affect the location of the minima),

\begin{align*}
&\mathcal{D}_{JSD}\left(p,q\right) = \left(\int p \log(\frac{p}{q}) dx + \int q \log(\frac{q}{p}) dx\right),\\
&\frac{d\mathcal{D}_{JSD}}{dq} = \left(\int{-\frac{p}{q}}dx + \int\log{\frac{q}{p}dx} + \int{q\frac{\frac{1}{p}}{\frac{q}{p}}dx}\right),\\
&\frac{d\mathcal{D}_{JSD}}{dq} = \left(\frac{d\mathcal{D}_{KL}}{dq} + \int\log{\frac{q}{p}dx} + \textrm{Cte} \right).
\label{eq:JSDLoss}
\end{align*}

\noindent Also, considering the changes between the prediction and ground truth is not significant (which is the case for most of the video frames), 
\begin{equation*}
q = p + \epsilon,\textrm{   where    } \epsilon \rightarrow 0,
\end{equation*}

\noindent the second term in $\frac{d\mathcal{D}_{JSD}}{dq}$ will be approximately zero, as follows, 
\begin{align*}
\int{\log{\frac{p+\epsilon}{p}}dx} = \int{\log{\frac{p+\epsilon}{p}}dx} = \int{\log{(1 + \frac{\epsilon}{p})}dx} \approx 0.
\end{align*}

\noindent As a result of this, the differentiation of both loss function will have the following relationship, 

\begin{align*}
&\frac{d\mathcal{D}_{JSD}}{dq} = \frac{d\mathcal{D}_{KL}}{dq} + \textrm{Cte}.
\end{align*}

\noindent The fact that the resulting differentiation signals for these loss functions are highly similar can explain why the generated tracking and filtering performance using these two objective functions were nearly identical.

\section{Evaluation metrics}
\subsection{Filtering metric definition}
\label{sec:OSPA}
For quantitative evaluation, we compute the Optimal Sub-Pattern Assignment (OSPA, \cite{Schuhmacher:2008}) distance, which is an improved version of the Optimal Mass Transfer (OMAT, \cite{Hoffman:2004}). OSPA distance has been extensively utilised for evaluating the accuracy of multi-target filtering algorithms~\cite{Beard:2017,Fantacci:2018,Meyer:2017,Vo:2017}. Assuming two sets $\mathcal{A}=\{m_1,m_2,\ldots,m_\alpha\}$ and $\mathcal{B}=\{n_1,n_2,\ldots,n_\beta\}$, the OSPA distance of order $p$ and cut-off $c$ is defined as \cite{Schuhmacher:2008},
\begin{equation}
\textrm{OSPA}(\mathcal{A},\mathcal{B}) = \frac{1}{\max\{\alpha,\beta\}} \bigg( c^p |\alpha-\beta| + cost \bigg)^{1/p},
\label{eq:OSPA}
\end{equation}
where $\alpha$ and $\beta$ are $\mathcal{A}$ and $\mathcal{B}$ cardinalities, respectively. The OSPA error consists of two terms: (1) cardinality error (Card), which computes the difference in the number of elements in the sets $X$ and $Y$; and (2) the localisation {\emph{cost}} (Loc), which is the smallest pair-wise distance among all the elements in the two sets (the best-worst error \cite{Emambakhsh:2017}), which in our work, is computed via the Hungarian assignment. Similar to \cite{Schuhmacher:2008}, we choose $p=1$ and $c=100$.

\subsection{Tracking metrics definition}
Depending on whether: (1) a target is a true positive (TP) (a bounding box detected correctly with intersection over union with the ground truth greater than 0.5), (2) false positive (FP) (wrong detection), or (3) false negative (FN) (missed detection), the total precision and recall per video can be computed as follows,
\begin{align}
&\textrm{Precision} = \frac{\sum_{i}TP_i}{\sum_{i} (TP_i+FP_i)} \\
&\textrm{Recall}=\frac{\sum_{i}TP_i}{\sum_{i} (TP_i+FN_i)},
\end{align}
\noindent which is obtained via adding the number of all positive and false negative samples over all frames.
The Multi Object Tracking Accuracy (MOTA) can be calculated as follows~\cite{Ristani:2016},
\begin{equation}\label{eq:MOTA}
\textrm{MOTA} = 1 - \frac{\sum_{i} (FN_i + FP_i + IdSW_i)}{\sum GT_i},
\end{equation}
where $IdSW_i$ is the number of ID switches among targets in frame $i$. $GT_i$ is the total number of ground truths in frame $i$.
The Multiple Object Tracking Accuracy with Logarithmic ID switches (MOTAL) metric is also defined as,
\begin{equation}
\textrm{MOTAL} = 1 - \frac{\sum_{i} FN_i + \sum_{i} FP_i + \log_{10} (\sum_{i} IdSW_i)}{\sum GT_i}.
\end{equation}

The implementation available at \url{https://bitbucket.org/amilan/motchallenge-devkit/} is used to evaluate these metrics.

\subsection{One time step at a glance}
\begin{figure*}[!t]
    \centering
    \begin{subfigure}[b]{0.35\textwidth}
        \includegraphics[width=\textwidth]{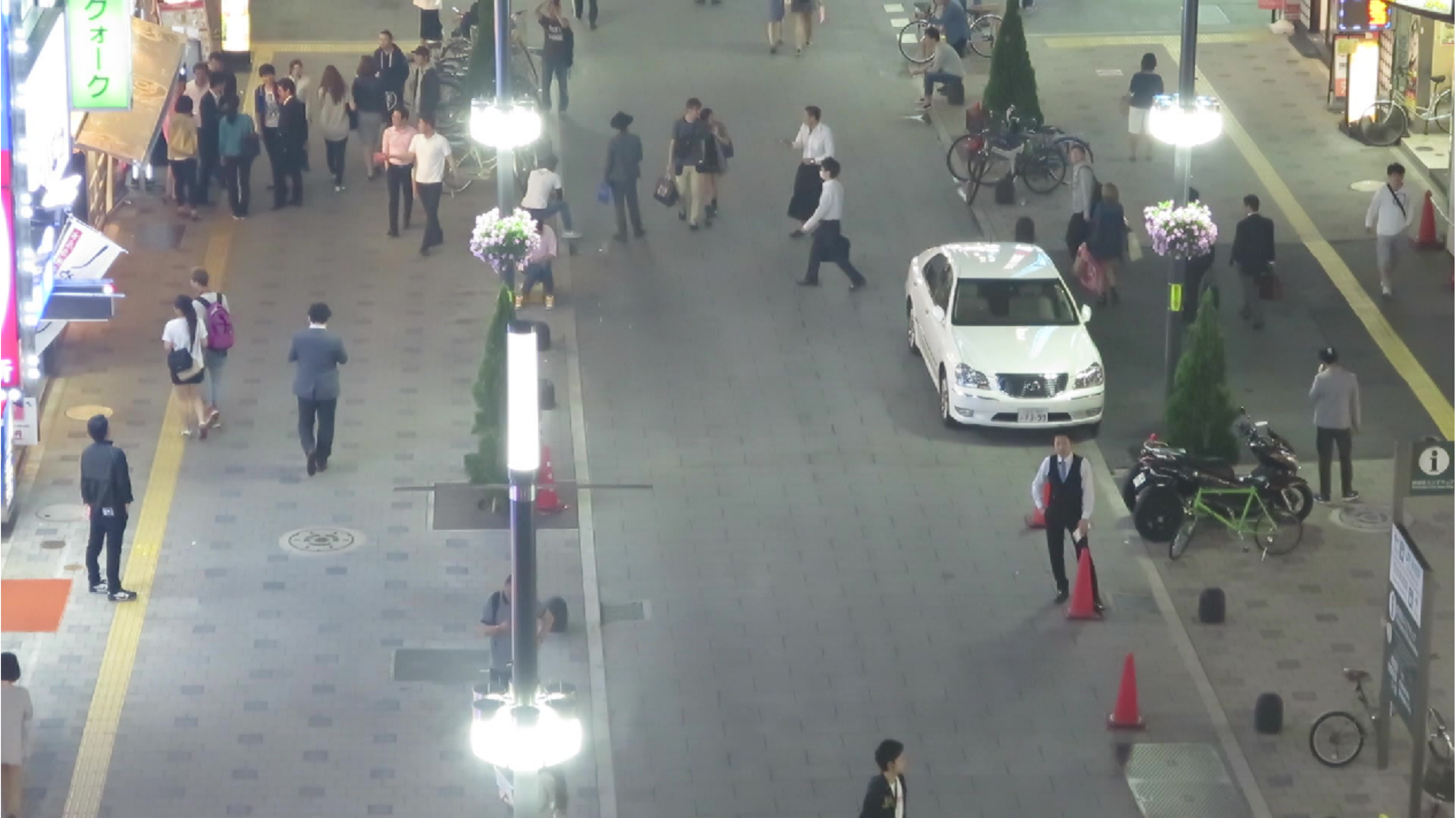}
        \caption{}
    \end{subfigure}
    \begin{subfigure}[b]{0.35\textwidth}
        \includegraphics[width=\textwidth]{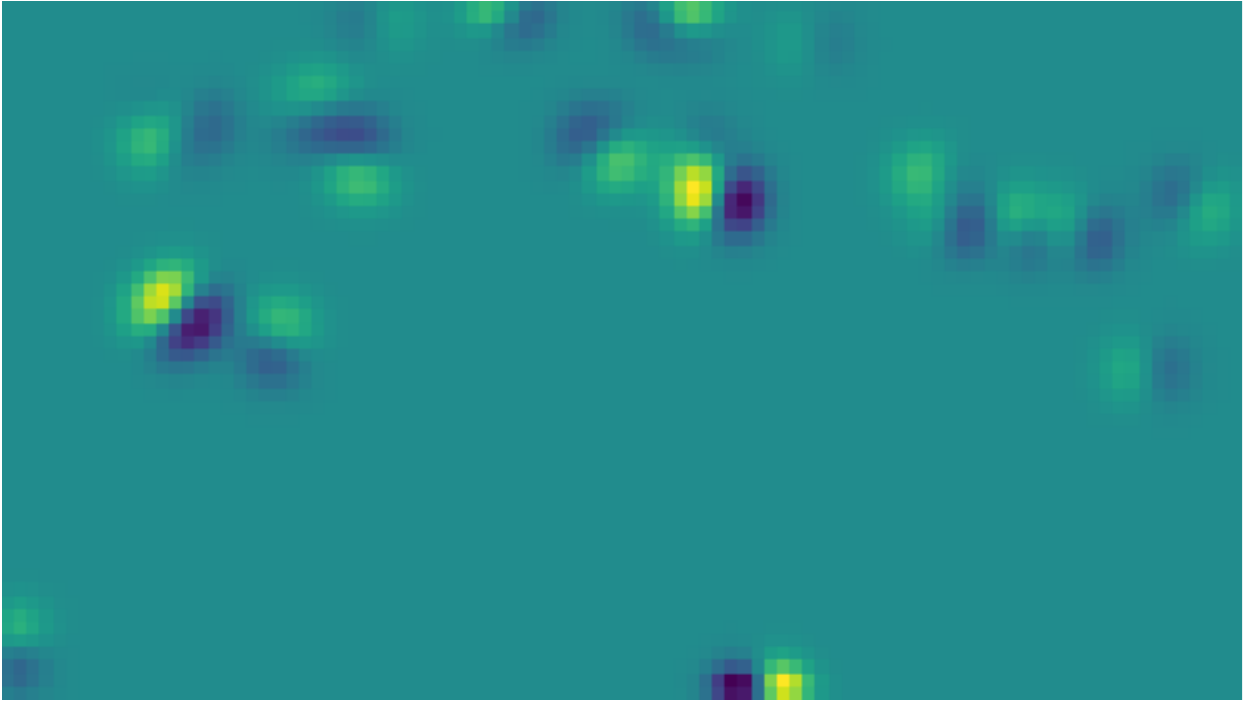}
        \caption{}
    \end{subfigure}
    \begin{subfigure}[b]{0.55\textwidth}
        \includegraphics[width=\textwidth]{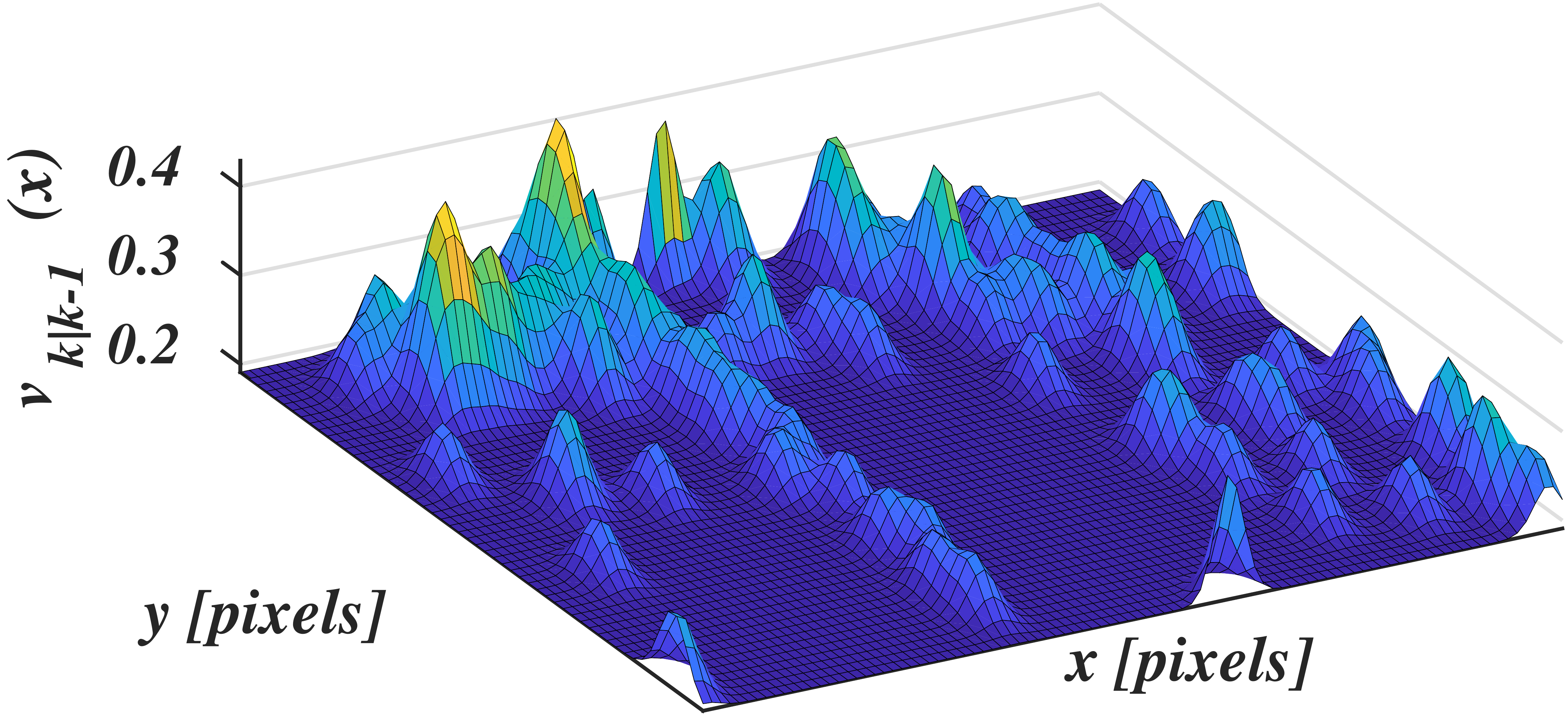}
        \caption{}
    \end{subfigure}
    \caption{Prediction step: (a) $D_{k, k-1}^{(i)}$; (b) Image at the $(k-1)^{th}$ iteration; (c) $v_{k|k-1}$}
    \label{fig:prediction}
\end{figure*}

In this section, we show some of the outputs of the important steps within the proposed MTFT pipeline.
Figure~\ref{fig:prediction}-b shows the output PDD map from the ConvLSTM prediction step, i.e $D_{k, k-1}^{(i)}$, for the current image at the $k-1^{th}$ time step, shown in Fig.~\ref{fig:prediction}-a. The peaks in $D_{k, k-1}^{(i)}$ indicate those regions corresponding to ``faster" target movement. Here by faster we mean how quickly a target is moving with respect to its covariance matrix. For such targets, the corresponding peak in the next time step's PHD function will be farther away, resulting in a high peak after subtraction, creating the bright yellow regions in Fig.~\ref{fig:prediction}-b. On the other hand, the darker regions indicate those targets which are mostly stationary.

The corresponding predicted PHD function ($v_{k|k-1}(x)$) is then calculated, which is shown in Fig.~\ref{fig:prediction}-c, where the peaks of $v_{k|k-1}(x)$ correspond to the expected location of targets. After obtaining the measurements at the $k^{th}$ time step ($Z_k$), the updated PHD function $v_k(x)$ is calculated. An overlaid plot of $v_k(x)$ and the $k^{th}$ image is illustrated in Fig.~\ref{fig:update}, where the peaks show the expected locations of the targets. This new PHD function is used to compute $D_{k+1, k}$, which is appended to the previous batch, to (online) train the ConvLSTM and predict for the next time step.
\begin{figure}[t]
\centering
\includegraphics[width=0.48\textwidth]{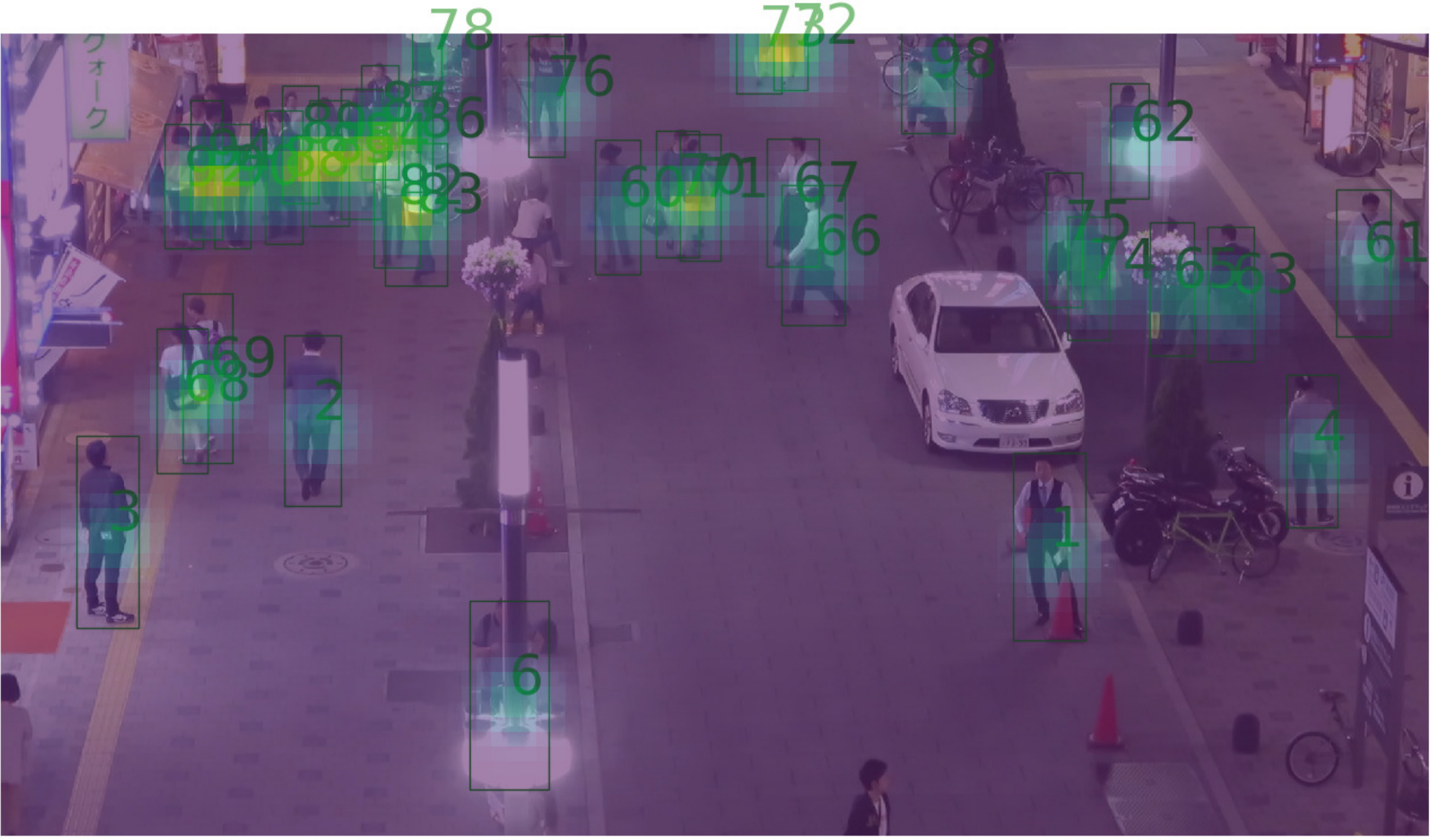}
\caption{Overlaid plot of $v_k(x)$ over the current image.}
\label{fig:update}
\end{figure}

\section{List of symbols}
{\footnotesize
\begin{itemize}
\item[] $k$: time step
\item[] $d$: state space dimensionality
\item[] $d_m$: measurement (observation) space dimensionality
\item[] $N$: input batch size of the neural network during training
\item[] $X_k$: RFS containing target state tuples
\item[] ${\bf{x}}^{(k)}_{k^\prime}$: ${k^\prime}^{th}$ target state tuple at time $k$
\item[] $M_k$: number of targets at time $k$
\item[] ${\bf{m}}^{(k)}_{k^\prime}$: mean vector of the ${k^\prime}^{th}$ target state tuple at time $k$
\item[] ${\bf{\Sigma}}^{(k)}_{k^\prime}$: covariance matrix of the ${k^\prime}^{th}$ target state tuple at time $k$
\item[] ${\bf{\omega}}^{(k)}_{k^\prime}$: Gaussian mixture weight of the ${k^\prime}^{th}$ target state tuple at time $k$
\item[] ${\bf{\mathcal{L}}}^{(k)}_{k^\prime}$: track label of the ${k^\prime}^{th}$ target state tuple at time $k$
\item[] ${\bf{a}}^{(k)}_{k^\prime}$: track age of the ${k^\prime}^{th}$ target state tuple at time $k$
\item[] ${\bf{\mathcal{M}}}^{(k)}_{k^\prime}$: motion vector of the ${k^\prime}^{th}$ target state tuple at time $k$
\item[] $v_k(x)$: probability hypothesis density map over target state space $x$ at time $k$
\item[] $D_{k, k-1}(x)$: probability density difference map over target state space $x$
\item[] $\mathcal{D}_{KL}(\bullet)$: Kullback-Leibler divergence loss
\item[] $D^{pred, (t)}_{k-i, k-i-1}(x)$: predicted target state at the $t^{th}$ epoch computed using the density functions at $(k-i)^{th}$ and $(k-i-1)^{th}$.
\item[] $D^{(i)}_{k, k-1}(x)$: initial predicted probability density difference map
\item[] $v^{(i)}_{k|k-1}(x)$: initial predicted probability hypothesis density map
\item[] ${\bf{m}}^{(k|k-1)}_{k^\prime}$: ${k^\prime}^{th}$ predicted mean state
\item[] $m_{k|k-1}$: predicted mean state RFS
\item[] $\omega^{(k|k-1)}_{k^\prime}$: ${k^\prime}^{th}$ predicted Gaussian mixture weight
\item[] $\Omega_{k|k-1}$: predicted Gaussian mixture weight RFS
\item[] $X^{(b)}_{k}$: target birth RFS
\item[] ${{\bf{x}}^{b}_{k^\prime}}^{(k)}$: ${k^\prime}^{th}$ birth target tuple at time $k$
\item[] $M^b_k$: number of target births at time $k$
\item[] ${{\bf{m}}^{b}_{k^\prime}}^{(k)}$: ${k^\prime}^{th}$ birth target mean state at time $k$
\item[] ${\bf{\Sigma}}_{birth}$: birth target covariance matrix
\item[] $\omega_{birth}$: birth target covariance matrix
\item[] ${\mathcal{L}}_{birth}$: birth target label
\item[] ${\mathcal{M}}_{birth}$: birth target motion vector
\item[] $X^{(+b)}_{k|k-1}$: final predicted target RFS
\item[] ${{\bf{m}}_{k^\prime}^{+b}}^{(k|k-1)}$: ${k^\prime}^{th}$ target's final (after appending births) predicted mean state
\item[] ${{\bf{\Sigma}}_{k^\prime}^{+b}}^{(k|k-1)}$: ${k^\prime}^{th}$ target's final (after appending births) predicted covariance matrix
\item[] ${{\omega}_{k^\prime}^{+b}}^{(k|k-1)}$: ${k^\prime}^{th}$ target's final (after appending births) predicted Gaussian mixture weight
\item[] ${\bf{H}}^{(k)}$: prediction to measurement space mapping matrix at time $k$
\item[] $\kappa_k(\bullet)$: clutter density at time $k$
\item[] $\bf{z}$: a vector over the measurement space 
\item[] $q^{(k)}_{k^{\prime\prime}}\left(\bullet\right)$: Gaussian distribution (likelihood) over the measurement space using the ${k^{\prime\prime}}^{th}$ target at time $k$
\item[] ${\bf{R}}^{(k)}$: covariance of the measurement noise at time $k$
\item[] ${\bf{K}}^{(k)}_{k^\prime}$: Kalman gain at time $k$ for the ${k^{\prime}}^{th}$ target
\item[] $X^{(s)}_k$: initial updated target RFS
\item[] $X^{(s)}_k$: initial unlabelled updated target RFS
\item[] ${{\bf{x}}_{k^\prime}^{(s)}}^{(k)}$: $k^{\prime}$ unlabelled updated target tuple
\item[] $\textrm{IoU}(\bullet, \bullet)$: intersection over union function
\item[] $\delta_{k^{\prime\prime}, k^{\prime}}^{(k)}$: distance (similarity) measurement between ${k^\prime}^{th}$ unlabelled updated target at time $k$ and ${k^{\prime\prime}}^{th}$ target at time $k-1$
\item[] $a_{am}$: target age amplification factor
\item[] $a_{at}$: target age attenuation factor
\item[] $a_{T}$: target age threshold
\item[] $a_{init}$: initial target age
\item[] $X^{(t2t)}_k$: track to track association RFS at time $k$
\item[] $M^{(t2t)}_k$: cardinality of $X^{(t2t)}_k$
\item[] $a^{(t2t)}_i$: age of the $i^{th}$ target in $X^{(t2t)}_k$ at time $k$
\item[] ${{\bf{x}}^{(t2t)}_i}^{(k)}$: $i^{th}$ target tuple in $X^{(t2t)}_k$ at time $k$
\end{itemize}
}

{\small
\bibliographystyle{ieee}
\bibliography{refs}

\begin{thebibliography}{10}\itemsep=-1pt

\bibitem{Bay2016}
A.~Bay, S.~Lepsoy, and E.~Magli.
\newblock Stable limit cycles in recurrent neural networks.
\newblock In {\em 2016 International Conference on Communications (COMM)},
  pages 89--92, 2016.

\bibitem{Beard:2017}
M.~Beard, B.~T. Vo, B.~N. Vo, and S.~Arulampalam.
\newblock Void probabilities and {C}auchy-{S}chwarz divergence for generalized
  labeled multi-{B}ernoulli models.
\newblock {\em IEEE Transactions on Signal Processing}, 65(19):5047--5061,
  2017.

\bibitem{Bewley:2016}
A.~Bewley, Z.~Ge, L.~Ott, F.~Ramos, and B.~Upcroft.
\newblock Simple online and realtime tracking.
\newblock In {\em International Conference on Image Processing (ICIP)}, pages
  3464--3468, 2016.

\bibitem{Dollar:2014}
P.~Doll{\'a}r, R.~Appel, S.~Belongie, and P.~Perona.
\newblock Fast feature pyramids for object detection.
\newblock {\em IEEE Transactions on Pattern Analysis and Machine Intelligence},
  36(8):1532--1545, 2014.

\bibitem{Emambakhsh:2019}
M.~Emambakhsh, A.~Bay, and E.~Vazquez.
\newblock Deep recurrent neural network for multi-target filtering.
\newblock In {\em International Conference on MultiMedia Modeling (MMM)}, pages
  519--531, 2019.

\bibitem{Emambakhsh:2010}
M.~Emambakhsh, H.~Ebrahimnezhad, and M.~Sedaaghi.
\newblock Integrated region-based segmentation using color components and
  texture features with prior shape knowledge.
\newblock {\em International Journal of Applied Mathematics and Computer
  Science}, 20(4):711--726, 2010.

\bibitem{Emambakhsh:2017}
M.~Emambakhsh and A.~Evans.
\newblock Nasal patches and curves for expression-robust {3D} face recognition.
\newblock {\em IEEE Transactions on Pattern Analysis and Machine Intelligence},
  39(5):995--1007, 2017.

\bibitem{Fantacci:2018}
C.~Fantacci, B.~N. Vo, B.~T. Vo, G.~Battistelli, and L.~Chisci.
\newblock Robust fusion for multisensor multiobject tracking.
\newblock {\em IEEE Signal Processing Letters}, 25(5):640--644, 2018.

\bibitem{Felzenszwalb:2010}
P.~F. Felzenszwalb, R.~B. Girshick, D.~McAllester, and D.~Ramanan.
\newblock Object detection with discriminatively trained part-based models.
\newblock {\em IEEE Transactions on Pattern Analysis and Machine Intelligence},
  32(9):1627--1645, 2010.

\bibitem{Gordon:2018}
D.~Gordon, A.~Farhadi, and D.~Fox.
\newblock Re$^3$: Real-time recurrent regression networks for visual tracking
  of generic objects.
\newblock {\em IEEE Robotics and Automation Letters}, 3(2):788--795, 2018.

\bibitem{Graves:2013}
A.~Graves, A.-r. Mohamed, and G.~Hinton.
\newblock Speech recognition with deep recurrent neural networks.
\newblock In {\em IEEE International Conference on Acoustics, Speech and Signal
  Processing (ICASSP)}, pages 6645--6649. IEEE, 2013.

\bibitem{Henriques:2015}
J.~F. Henriques, R.~Caseiro, P.~Martins, and J.~Batista.
\newblock High-speed tracking with kernelized correlation filters.
\newblock {\em IEEE Transactions on Pattern Analysis and Machine Intelligence},
  37(3):583--596, 2015.

\bibitem{hochreiter1997}
S.~Hochreiter and J.~Schmidhuber.
\newblock Long short-term memory.
\newblock {\em Neural Computation}, 9(8):1735--1780, 1997.

\bibitem{Hoffman:2004}
J.~R. Hoffman and R.~P.~S. Mahler.
\newblock Multitarget miss distance via optimal assignment.
\newblock {\em IEEE Transactions on Systems, Man, and Cybernetics - Part A:
  Systems and Humans}, 34(3):327--336, 2004.

\bibitem{Krizhevsky:2012}
A.~Krizhevsky, I.~Sutskever, and G.~E. Hinton.
\newblock Imagenet classification with deep convolutional neural networks.
\newblock In {\em Advances in Neural Information Processing Systems}, pages
  1097--1105, 2012.

\bibitem{leal:2015}
L.~Leal-Taix{\'e}, A.~Milan, I.~Reid, S.~Roth, and K.~Schindler.
\newblock {MOTChallenge} 2015: Towards a benchmark for multi-target tracking.
\newblock {\em arXiv preprint arXiv:1504.01942}, 2015.

\bibitem{Mahler:2007}
R.~Mahler.
\newblock {PHD} filters of higher order in target number.
\newblock {\em IEEE Transactions on Aerospace and Electronic Systems},
  43(4):1523--1543, 2007.

\bibitem{Mahler:2003}
R.~P.~S. Mahler.
\newblock Multitarget {Bayes} filtering via first-order multitarget moments.
\newblock {\em IEEE Transactions on Aerospace and Electronic Systems},
  39(4):1152--1178, 2003.

\bibitem{Meyer:2017}
F.~Meyer, P.~Braca, P.~Willett, and F.~Hlawatsch.
\newblock A scalable algorithm for tracking an unknown number of targets using
  multiple sensors.
\newblock {\em IEEE Transactions on Signal Processing}, 65(13):3478--3493,
  2017.

\bibitem{Milan:2017}
A.~Milan, S.~H. Rezatofighi, A.~Dick, I.~Reid, and K.~Schindler.
\newblock Online multi-target tracking using recurrent neural networks.
\newblock In {\em Thirty-First AAAI Conference on Artificial Intelligence},
  2017.

\bibitem{Moratuwage:2014}
D.~Moratuwage, D.~Wang, A.~Rao, N.~Senarathne, and H.~Wang.
\newblock {RFS} collaborative multivehicle {SLAM}: {SLAM} in dynamic
  high-clutter environments.
\newblock {\em IEEE Robotics Automation Magazine}, 21(2):53--59, 2014.

\bibitem{Nagappa:2017}
S.~Nagappa, E.~D. Delande, D.~E. Clark, and J.~Houssineau.
\newblock A tractable forward-backward {CPHD} smoother.
\newblock {\em IEEE Transactions on Aerospace and Electronic Systems},
  53(1):201--217, 2017.

\bibitem{Ren:2015}
S.~Ren, K.~He, R.~Girshick, and J.~Sun.
\newblock Faster {R-CNN}: Towards real-time object detection with region
  proposal networks.
\newblock In {\em Advances in Neural Information Processing Systems}, pages
  91--99, 2015.

\bibitem{Reuter:2014}
S.~Reuter, B.~T. Vo, B.~N. Vo, and K.~Dietmayer.
\newblock The labeled multi-{B}ernoulli filter.
\newblock {\em IEEE Transactions on Signal Processing}, 62(12):3246--3260,
  2014.

\bibitem{Ristani:2016}
E.~Ristani, F.~Solera, R.~Zou, R.~Cucchiara, and C.~Tomasi.
\newblock Performance measures and a data set for multi-target, multi-camera
  tracking.
\newblock In {\em European Conference on Computer Vision (ECCV)}, pages 17--35.
  Springer, 2016.

\bibitem{Schuhmacher:2008}
D.~Schuhmacher, B.~T. Vo, and B.~N. Vo.
\newblock A consistent metric for performance evaluation of multi-object
  filters.
\newblock {\em IEEE Transactions on Signal Processing}, 56(8):3447--3457, 2008.

\bibitem{xingjian2015convlstm}
X.~Shi, Z.~Chen, H.~Wang, D.-Y. Yeung, W.-K. Wong, and W.-c. Woo.
\newblock Convolutional {LSTM} network: A machine learning approach for
  precipitation nowcasting.
\newblock In {\em Advances in Neural Information Processing Systems}, pages
  802--810, 2015.

\bibitem{Shu:2012}
G.~Shu, A.~Dehghan, O.~Oreifej, E.~Hand, and M.~Shah.
\newblock Part-based multiple-person tracking with partial occlusion handling.
\newblock In {\em IEEE Conference on Computer Vision and Pattern Recognition
  (CVPR)}, pages 1815--1821. IEEE, 2012.

\bibitem{Vo:2006}
B.~N. Vo and W.~K. Ma.
\newblock The {G}aussian mixture probability hypothesis density filter.
\newblock {\em IEEE Transactions on Signal Processing}, 54(11):4091--4104,
  2006.

\bibitem{Vo:2005}
B.~N. Vo, S.~Singh, and A.~Doucet.
\newblock Sequential {M}onte {C}arlo methods for multitarget filtering with
  random finite sets.
\newblock {\em IEEE Transactions on Aerospace and Electronic Systems},
  41(4):1224--1245, 2005.

\bibitem{Vo:2017}
B.~N. Vo, B.~T. Vo, and H.~G. Hoang.
\newblock An efficient implementation of the generalized labeled
  multi-{B}ernoulli filter.
\newblock {\em IEEE Transactions on Signal Processing}, 65(8):1975--1987, 2017.

\bibitem{Vo:2014}
B.~N. Vo, B.~T. Vo, and D.~Phung.
\newblock Labeled random finite sets and the {Bayes} multi-target tracking
  filter.
\newblock {\em IEEE Transactions on Signal Processing}, 62(24):6554--6567,
  2014.

\bibitem{Weinzaepfel:2013}
P.~Weinzaepfel, J.~Revaud, Z.~Harchaoui, and C.~Schmid.
\newblock {D}eep{F}low: Large displacement optical flow with deep matching.
\newblock In {\em International Conference on Computer Vision}, pages
  1385--1392, 2013.

\bibitem{wojke2017simple}
N.~Wojke, A.~Bewley, and D.~Paulus.
\newblock Simple online and realtime tracking with a deep association metric.
\newblock In {\em International Conference on Image Processing (ICIP)}, pages
  3645--3649, 2017.

\bibitem{Yang:2016}
F.~Yang, W.~Choi, and Y.~Lin.
\newblock Exploit all the layers: Fast and accurate cnn object detector with
  scale dependent pooling and cascaded rejection classifiers.
\newblock In {\em Proceedings of the IEEE Conference on Computer Vision and
  Pattern Recognition}, pages 2129--2137, 2016.

\end{thebibliography}
}

\end{document}